\documentclass[runningheads]{llncs}

\usepackage{eccv}

\newcommand{\x}{\ensuremath{\boldsymbol{x}}}
\newcommand{\xw}{\ensuremath{\boldsymbol{x}^w}}
\newcommand{\xl}{\ensuremath{\boldsymbol{x}^l}}

\newcommand{\vc}{\ensuremath{\boldsymbol{c}}}
\newcommand{\I}{\ensuremath{\boldsymbol{I}}}
\newcommand{\bbE}{\ensuremath{\mathbb{E}}}
\newcommand{\bbD}{\ensuremath{\mathbb{D}}}
\newcommand{\kl}{\ensuremath{\bbD_{\text{KL}}}}

\newcommand{\calvar}[1]{\ensuremath{\mathcal{#1}}}
\newcommand{\calD}{\calvar{D}}
\newcommand{\calN}{\calvar{N}}

\newcommand{\noise}{\ensuremath{\boldsymbol{\epsilon}}}

\newcommand{\pref}{p_{\text{ref}}}

\newcommand{\dohyun}[1]{\textcolor{Orange}{[DH: #1]}}
\newcommand{\previous}[1]{\textcolor{red}{[Prev: #1]}}

\renewcommand{\dohyun}[1]{#1}
\renewcommand{\previous}[1]{}

\usepackage{eccvabbrv}

\usepackage{amsmath}
\usepackage{amssymb}
\usepackage{bm} 

\usepackage[table,dvipsnames]{xcolor}  
\usepackage{mathtools} 
\usepackage{subcaption}                
\usepackage{pifont}                    

\usepackage{tikz}
\usetikzlibrary{shapes.geometric}
\definecolor{bkcolor}{HTML}{1F77B4}
\definecolor{ourscolor}{HTML}{FF7F0E}
\usepackage{listings}

\usepackage{graphicx}
\usepackage{booktabs}

\usepackage[accsupp]{axessibility}  

\usepackage{hyperref}

\usepackage{orcidlink}

\begin{document}

\title{Direct Diffusion Score Preference Optimization via Stepwise Contrastive Policy-Pair Supervision} 

\titlerunning{Direct Diffusion Score Preference Optimization}

\author{Dohyun Kim\inst{1} \and
Seungwoo Lyu\inst{1} \and
Seung Wook Kim\inst{2} \and
Paul Hongsuck Seo\inst{1}}

\authorrunning{D.~Kim et al.}

\institute{$^{1}$Dept. of CSE, Korea University \quad $^{2}$NVIDIA\\
\email{\{a12s12, dbtmddn41, phseo\}@korea.ac.kr, seungwookk@nvidia.com}}

\maketitle

\begin{abstract}
Diffusion models have achieved impressive results in generative tasks such as text-to-image synthesis, yet they often struggle to fully align outputs with nuanced user intent and maintain consistent aesthetic quality. 
Existing preference-based training methods such as Diffusion Direct Preference Optimization help address these issues, but obtain their supervision targets from the forward process $q(\x_{t-1}\!\mid\!\x_t,\x_0)$ derived from terminal samples, which is not directly aligned with the model’s actual backward denoising transitions at each step. In this work, we introduce Direct Diffusion Score Preference Optimization (DDSPO), which defines stepwise preference supervision directly over backward denoising transitions through a contrastive policy pair, rather than relying on forward-process approximations from terminal samples. We propose two practical instantiations of the contrastive policy pair: training separate winning and losing models on preference data, and inducing a contrastive policy pair without additional training by using a pretrained reference model conditioned on an original prompt and a semantically degraded variant, requiring neither reward modeling nor manual annotations. \dohyun{Empirical results show that contrastive-policy-pair supervision is more effective than forward-process-based supervision across text--image alignment and aesthetic-quality tasks.}
\previous{Empirical results across multiple tasks demonstrate the effectiveness of stepwise contrastive supervision, with DDSPO consistently improving text--image alignment and aesthetic quality.}
Our implementation is available at: \url{https://dohyun-as.github.io/DDSPO}
\end{abstract}
  
\section{Introduction}
\label{sec:intro}
Diffusion models~\cite{ddpm,ldm} have achieved impressive results across a variety of generative tasks, particularly in text-to-image synthesis~\cite{sdm_v1.5,SDXL,Imagen,dalle2}.
Despite this progress, they often struggle to fully align generated outputs with nuanced user intent and to consistently produce aesthetically high-quality images.
Addressing these shortcomings typically requires task-specific training data, which can be difficult and costly to obtain. 
As an alternative, human preference annotations provide a lightweight yet expressive means of encoding such qualitative information in the form of ranked comparisons. 
Leveraging this, recent approaches have incorporated human preferences into the training process, such as through reinforcement learning with human feedback (RLHF~\cite{rlhf,rlhfonlm}). 
A notable example is Diffusion DPO~\cite{Wallace_2024_CVPR}, which extends the Direct Preference Optimization (DPO) framework~\cite{rafailov2023direct}—originally developed for language and vision-language models~\cite{rafailov2023direct, xing2025re, xie2024v}—to diffusion models. 
Diffusion DPO enables models to learn directly from pairwise preferences over final samples without requiring reward models.
However, it approximates backward denoising transitions using the forward process, which does not reflect the model’s actual reverse denoising process at inference time, and thus fails to provide stepwise supervision aligned with the model’s denoising trajectories.

To overcome these limitations, we propose \textbf{Direct Diffusion Score Preference Optimization (DDSPO)} that formulates preference supervision in score space at each diffusion timestep rather than in the final sample space. 
Unlike prior approaches that define preferences over final samples \( (\x_0^w, \x_0^l) \), DDSPO introduces stepwise supervision by comparing the denoising scores of a contrastive policy pair at each timestep.
This is implemented by extending preference labels to tuples  \(((\x_t^w, \x_{t-1}^w), (\x_t^l, \x_{t-1}^l))\),
derived from the contrastive policy pair, \ie, preferred and dispreferred denoising policies.
We define a score-space loss that optimizes the model’s denoising predictions toward targets \( \boldsymbol{\epsilon}_\star^w \) and \( \boldsymbol{\epsilon}_\star^l \), corresponding to preferred and dispreferred behavior at each timestep. 
By deriving per-timestep targets from the contrastive policy pair,
DDSPO enables richer supervision across denoising steps and reduces reliance on final outputs \( \x_0 \).
To instantiate the contrastive policy pair, we propose two complementary approaches: a Data-Driven Contrastive Policy Pair (DD-CPP), which trains separate models on preferred and dispreferred samples, and a Training-Free Contrastive Policy Pair (TF-CPP), which directly constructs stepwise contrastive directions from a pretrained reference model using an original prompt $\vc$ and a semantically degraded variant $\vc^{-}$.
\dohyun{As a result, DDSPO consistently improves its base models and remains competitive with human-supervised methods across text--image alignment and aesthetic quality.
More importantly, controlled comparisons show that stepwise supervision derived from contrastive policy pairs is more effective than supervision based on forward-process approximations.}
\previous{As a result, DDSPO achieves competitive or superior performance across text--image alignment and aesthetic quality, demonstrating that stepwise contrastive supervision over backward denoising transitions is more effective than forward-process-based approximations.}
Our main contributions are threefold: 
\begin{itemize}
    \item We propose DDSPO, a preference optimization framework that enables direct supervision over intermediate denoising steps in score space through a contrastive policy pair.
    
    \item
    We introduce two practical instantiations of the contrastive policy pair: DD-CPP, which learns winning/losing policies from preference-labeled data, and TF-CPP, which constructs stepwise contrastive directions from a pretrained reference model using an original prompt $\vc$ and a semantically degraded variant $\vc^{-}$, without preference-labeled data.

    \item 
    Extensive experiments demonstrate that stepwise supervision derived from a contrastive policy pair is effective, yielding consistent improvements in text--image alignment and aesthetic quality over diffusion DPO--style baselines.
\end{itemize}

\section{Related work}
\label{sec:rel_work}

\noindent\textbf{Improving Diffusion Model} \ \ 
Several lines of research~\cite{pickscore, xu2023imagereward, black2024trainingdiffusionmodelsreinforcement, lee2025calibrated, fan2023dpok, yang2024usinghumanfeedbackfinetune} have sought to improve diffusion models by incorporating external signals derived from human preferences. These approaches often rely on pretrained or learned reward models~\cite{pickscore, xu2023imagereward, black2024trainingdiffusionmodelsreinforcement, lee2025calibrated}, and formulate the generation process as a reinforcement learning (RL) problem~\cite{yang2024usinghumanfeedbackfinetune, black2024trainingdiffusionmodelsreinforcement, fan2023dpok, gu2024diffusion}. 
More recent work also explores combining multiple reward models or iteratively refining preferences~\cite{zhang2025itercomp, lee2025calibrated,  zhao2025fine, clark2023directly, hao2023optimizing, prabhudesai2023aligning}.

\noindent\textbf{Direct Preference Optimization (DPO)}
Direct Preference Optimization (DPO)~\cite{rafailov2023direct} has recently gained traction as a more direct and efficient way to align generative models with human preferences. Originally proposed for aligning large language models and vision-language models, DPO fine-tunes models using only paired preference data, bypassing the need for explicitly trained reward models~\cite{xing2025re, liu2024mia}. This leads to simpler training pipelines and improved alignment stability compared to reinforcement learning from human feedback. 
Diffusion DPO~\cite{Wallace_2024_CVPR} extends DPO to diffusion models 
by re-formulating the objective to match the diffusion sampling process. Several 
subsequent works~\cite{croitoru2024curriculum, zhu2025dspo, zhang2024aligning, liang2025aestheticposttrainingdiffusionmodels,hong2025marginaware,diffkto} extend this paradigm in different ways, exploring 
alternative objectives and forms of preference supervision.

\noindent\textbf{Self-Training}
Recent advances have shown that generative models can be effectively aligned with human preferences through self-training, without relying on paired preference data or externally trained reward models~\cite{zhu2024self, lee2025aligning, yuan2024self, chen2024learning, majumder2024tango, deng2024enhancing, he2019revisiting, xie2020self, wei2020theoretical, zoph2020rethinking, sohn2020fixmatch, ghiasi2021multi, kang2023dialog}.
Especially for diffusion models, these methods typically involve generating synthetic preference signals from the model’s own outputs—contrasting higher-quality generations with intentionally degraded ones~\cite{majumder2024tango} or leveraging iterative refinements across model checkpoints~\cite{yuan2024self}.
By treating the model’s stronger outputs as preferred examples and weaker ones as negatives, they enable preference-aware fine-tuning through supervised objectives such as Direct Preference Optimization~\cite{majumder2024tango, yuan2024self}.
In some cases, contrastive learning~\cite{lee2025aligning} or prompt relabeling~\cite{chen2024learning} further reinforces semantic consistency and corrects misalignments. 
DDSPO follows this direction but departs in how the synthetic contrast is used: rather than using degraded counterparts only to form terminal sample pairs, DDSPO instantiates the contrastive policy pair itself from the preferred/degraded conditions and derives per-timestep score targets from backward denoising transitions, enabling stepwise supervision that is aligned with the reverse denoising process.
\section{Method}
\label{sec:method}
Our objective is to propose a novel optimization formulation for aligning pretrained diffusion models with desirable generation behavior, such as improved text–to-image alignment or aesthetic quality.
We first review Direct Preference Optimization (DPO)~\cite{rafailov2023direct} and its extension to diffusion models in \cref{sec:prelim}, which assume access to preference-labeled final samples \((\x_0^w, \x_0^l \mid \vc)\).
In \cref{sec:method:DDSPO}, we introduce Direct Diffusion Score Preference Optimization (DDSPO), which directly optimizes denoising behavior at each timestep using winning and losing denoising scores sampled from a contrastive policy pair as stepwise supervision.
\cref{sec:construct_score} describes practical ways to instantiate the contrastive policy pair: a Data-Driven Contrastive Policy Pair (DD-CPP), which trains separate winning and losing models on preference-labeled data, and a Training-Free Contrastive Policy Pair (TF-CPP), which constructs stepwise contrastive directions from a pretrained reference model without preference-labeled data, using an original prompt $\vc$ and a semantically degraded variant $\vc^{-}$.
Finally, \cref{sec:method_prev_work} compares DDSPO with existing diffusion preference-optimization methods, including Diffusion DPO~\cite{Wallace_2024_CVPR}, DSPO~\cite{zhu2025dspo}, and SPO~\cite{liang2025aestheticposttrainingdiffusionmodels}, and discusses their conceptual and methodological differences in how preference signals are defined and utilized.

\subsection{Preliminary}
\label{sec:prelim}

\noindent\textbf{Direct Preference Optimization}
Direct Preference Optimization (DPO)~\cite{rafailov2023direct} is a learning framework that aligns model outputs with human preferences by directly optimizing on pairwise preference data.
Given a conditioning input \( \vc \), we assume access only to preference-labeled pairs \( (\x_0^w, \x_0^l \mid \vc) \), 
where $\x_0^w$ denotes the preferred sample and $\x_0^l$ the dispreferred one.
The interpretation of \( \vc \) and \( \x_0 \) depends on the task, where \( \vc \) is a text caption and \( \x_0 \) is a synthesized image in text-to-image generation. Such preferences may reflect various notions of quality—such as semantic alignment, aesthetic appeal, or factual correctness—depending on the task context.

These preferences can be formalized using the Bradley–Terry model~\cite{BT-model}, which defines the following distribution:
\begin{equation}
\mathbb{P}(\x_0^w \succ \x_0^l \mid \vc) = \sigma\left(r(\vc, \x_0^w) - r(\vc, \x_0^l)\right),
\label{equ:BT_distribution}
\end{equation}
where $\x_0^w \succ \x_0^l$ denotes that $\x_0^w$ is preferred over $\x_0^l$, $\sigma(\cdot)$ is the sigmoid function and $r(\vc, \x_0)$ is a latent reward function that is difficult to access directly. 
DPO reparameterizes this setup to directly optimize the model distribution $p_\theta(\x_0 \mid \vc)$. 
It begins from the RLHF-style KL-regularized objective, which maximizes reward while constraining the learned distribution to stay close to a fixed reference distribution \(p_{\text{ref}}(\x_0 \mid \vc)\):
\begin{equation}
\max_{\theta} \, \mathbb{E}_{\x_0 \sim p_\theta(\x_0 \mid \vc)} \left[r(\vc, \x_0)\right] - \beta \, \kl\left(p_\theta(\x_0 \mid \vc) \,\|\, p_{\text{ref}}(\x_0 \mid \vc)\right)
\label{equ:RLHF}
\end{equation}
where $\beta$ is a parameter that controls regularization strength.
This objective admits a closed-form solution where the optimal distribution \( p^*_\theta(\x_0 \mid \vc) \) is proportional to the reference distribution \( p_{\text{ref}}(\x_0 \mid \vc) \) scaled by an exponential of the reward: 
\(
p^*_\theta(\x_0 \mid \vc) \propto p_{\text{ref}}(\x_0 \mid \vc) \cdot \exp(r(\vc, \x_0)/\beta)
\). 
This reformulation reveals that the reward function can be captured by the ratio between the optimized policy and the reference distribution, eliminating the need to model \( r(\vc, x_0) \) explicitly. 
This leads to the DPO training objective:
\begin{equation}
\mathcal{L}_{\text{DPO}}(\theta) = - \mathbb{E}_{\vc, \x_0^w, \x_0^l} \log \sigma\bigg( 
 \beta \log \frac{p_\theta(\x_0^w \mid \vc)}{p_{\text{ref}}(\x_0^w \mid \vc)} -
\beta \log \frac{p_\theta(\x_0^l \mid \vc)}{p_{\text{ref}}(\x_0^l \mid \vc)}
\bigg).
\label{equ:DPO_loss}
\end{equation}

\noindent\textbf{DPO for Diffusion Models}
Applying DPO to diffusion models introduces a unique challenge: directly computing the log-likelihood ratio \( \log \tfrac{p_\theta(x_0 \mid \vc)}{p_{\text{ref}}(x_0 \mid \vc)} \) is intractable due to the need to marginalize over all possible diffusion trajectories \( x_{1:T} \) that generate \( x_0 \). To circumvent this, \cite{Wallace_2024_CVPR} reformulates the objective over entire denoising paths and approximate the reverse process \( p_\theta(x_{1:T} \mid x_0) \) using the forward noising process \( q(x_{1:T} \mid x_0) \). 
This reparameterization enables training in the \textit{score space}, where models learn to predict the noise (or equivalently, the score function~\cite{ddpm,song2021scorebased}) that guides the denoising process: 
\begin{align*}
\mathcal{L}(\theta) = 
& - \mathbb{E}_{\substack{(\x_0^w, \x_0^l) \sim \mathcal{D}, t \sim \mathcal{U}(0, T), \\ \x_t^w \sim q(\x_t^w|\x_0^w), \x_t^l \sim q(\x_t^l|\x_0^l) }} \log \sigma \bigg( - \beta \cdot \big[ \nonumber \\
&
\quad \| \noise^w - \noise_\theta(\x_t^w, t, \vc) \|_2^2 
- \| \noise^w - \noise_{\text{ref}}(\x_t^w, t, \vc) \|_2^2 \\ 
&- (\| \noise^l - \noise_\theta(\x_t^l, t, \vc) \|_2^2 
- \| \noise^l - \noise_{\text{ref}}(\x_t^l, t, \vc) \|_2^2)
\big] \bigg)
\end{align*}
where $\x_t^\ast = \alpha_t \x_0^\ast + \sigma_t \noise^\ast$ 
is drawn from $q(\x_t^\ast \mid \x_0^\ast)$, 
and $\noise^\ast \sim \mathcal{N}(0, I)$.
The term 
$\| \noise^w - \noise_\theta(\x_t^w, t, \vc) \|_2^2
- \| \noise^w - \noise_{\text{ref}}(\x_t^w, t, \vc) \|_2^2$
penalizes the student whenever its predicted score at $\x_t^w$ is farther 
from the target $\noise^w$ than the reference model, thereby pushing it 
closer to $\noise^w$.  
The analogous term for $\x_t^l$ pushes the student away from $\noise^l$ 
relative to the reference, producing a pairwise preference signal.
However, both $\noise^w$ and $\noise^l$ originate from 
forward-process samples $\x_t^\ast \sim q(\x_t^\ast \mid \x_0^\ast)$ and therefore ultimately depend on the terminal samples $(\x_0^w,\x_0^l)$, which do not reflect the model’s actual reverse denoising trajectories and thus provide only approximate supervision.
Full derivations of both the standard and diffusion DPO objectives are provided in the \cref{sec:supp_derivation}.

\subsection{Direct Diffusion Score Preference Optimization} 
\label{sec:method:DDSPO}
We now consider a setting in which preference is defined 
not over final generated samples, but directly over 
backward denoising transitions at intermediate steps, 
thereby grounding supervision in the actual reverse denoising process rather than using the forward process posterior $q(\x_{t-1}\mid \x_t,\x_0)$ as a surrogate supervision signal for reverse transitions.
Crucially, rather than assuming access only to preference-labeled final samples \( (\x_0^w, \x_0^l \mid \vc) \), we also assume access to preference-labeled denoising transitions at intermediate steps. 
Let \( p_\star^w(\x_{t-1,t} \mid \vc) \) and \( p_\star^l(\x_{t-1,t} \mid \vc) \) denote the winning and losing denoising policies, forming a contrastive policy pair.
Sampling from these policies yields tuples $((\x_t^w, \x_{t-1}^w), (\x_t^l, \x_{t-1}^l) \mid \vc)$.
This enables direct preference of denoising behavior at each timestep of the diffusion process.

Following the Bradley–Terry formulation in \cref{equ:BT_distribution}, we extend the preference supervision from final outputs \( (\x_0^w, \x_0^l \mid \vc) \) to denoising transitions at intermediate steps, using tuples of the form \( ((\x_t^w, \x_{t-1}^w), (\x_t^l, \x_{t-1}^l) \mid \vc) \).  
The reward function is accordingly redefined as \( r(\vc, \x_t, \x_{t-1}) \), and preferences are modeled over entire denoising transitions.
By applying the same derivation steps as in  \cref{equ:BT_distribution,equ:RLHF,equ:DPO_loss}, and extending the supervision to denoising transitions at arbitrary diffusion timesteps \( t \sim \mathcal{U}(0, T) \) the loss can be analogously formulated as:
\begin{multline*}
\mathcal{L}(\theta) = 
 - \mathbb{E}_{\vc\sim \mathcal{D}(c),\,t \sim \mathcal{U}(0, T), \,
(\x_{t-1}^{w}, \x_t^{w})
,\,
(\x_{t-1}^l, \x_t^l)
} \\
\log\sigma\!\left(
\beta \log\!\frac{p_{\theta}(\x_{t-1}^w\mid\x_t^w,\vc)}{p_{\text{ref}}(\x_{t-1}^w\mid\x_t^w,\vc)}
-\beta \log\!\frac{p_{\theta}(\x_{t-1}^l\mid\x_t^l,\vc)}{p_{\text{ref}}(\x_{t-1}^l\mid\x_t^l,\vc)}
\right),
\end{multline*}
where  $(\x_{t-1}^{w}, \x_t^{w}) \sim p^{w}_\star(\x_{t-1,t}^{w} \mid \vc) ,\, (\x_{t-1}^l, \x_t^l) \sim p^l_\star(\x_{t-1,t}^l \mid \vc)$.
Since the joint transition distribution \( p^w_\star(\x_{t-1,t}^w \mid \vc) \) is generally intractable, we approximate it as  
\( p^w_\star(\x_{t-1,t}^w \mid \vc) \approx q(\x_t^w \mid \x_0^w) \, p^w_\star(\x_{t-1}^w \mid \x_t^w, \vc) \),  
where \( \x_0^w \sim \mathcal{D} \), \( \x_t^w \sim q(\x_t^w \mid \x_0^w) \), and \( \x_{t-1}^w \sim p^w_\star(\x_{t-1}^w \mid \x_t^w, \vc) \).  
This provides a practical sampling scheme in which the preferred transition is constructed by forward noising followed by preference-guided denoising.  
An analogous approximation is applied to the dispreferred transition \( p^l_\star(\x_{t-1}^l, \x_t^l \mid \vc) \).
Under this approximation, we derive the following score-space objective~\cite{song2021scorebased}; see \cref{sec:supp_derivation} for the full derivation:
\begin{multline}
\label{eq:DDSPO}
\mathcal{L}_{\text{DDSPO}}(\theta) = 
 - \mathbb{E}_{\substack{(\x_0^w,\x_0^l) \sim \mathcal{D},\,\vc\sim \mathcal{D}(c), \, t \sim \mathcal{U}(0, T),\, \\ \x_t^w \sim q(\x_t \mid \x_0^w), \,  \x_t^l \sim q(\x_t \mid \x_0^l)}} \\
\log \sigma \bigg( - \beta \cdot \big[\| \noise_{\star}^w - \noise_\theta(\x_t^w, t, \vc) \|_2^2 
- \| \noise_{\star}^w - \noise_{\text{ref}}(\x_t^w, t, \vc) \|_2^2 \\
-\big( 
\| \noise_{\star}^l - \noise_\theta(\x_t^l, t, \vc) \|_2^2 
- \| \noise_{\star}^l - \noise_{\text{ref}}(\x_t^l, t, \vc) \|_2^2 
\big) 
\big] \bigg).
\end{multline}
Here, \( \noise^w_\star \), \( \noise^l_\star \) denote denoising score targets obtained from the contrastive policy pair---$p^w_\star(\x_{t-1}^w \mid \x_t^w, \vc)$ and $p^l_\star(\x_{t-1}^l \mid \x_t^l, \vc)$---thereby retaining 
the actual backward denoising transitions, in contrast to 
diffusion DPO which approximates these transitions 
via the forward process $q(\x_{t-1} \mid \x_t, \x_0)$, 
resulting in a mismatch between the supervision signal 
and the backward denoising process.

\subsection{Constructing Contrastive Policy Pair}
\label{sec:construct_score}

We consider two instantiations: (i) training a contrastive policy pair from preference-labeled image pairs (DD-CPP), and (ii) constructing step-wise contrastive directions from a pretrained reference model without additional training or preference-labeled data (TF-CPP).

\noindent \textbf{Data-Driven Contrastive Policy Pair (DD-CPP)} \ \
Given a preference-labeled dataset $\mathcal{D}=\{(\x_0^w,\x_0^l,\vc)\}$, we obtain a contrastive policy pair by training two separate models initialized from the pretrained diffusion model, which keeps their reverse denoising behavior close to the pretrained reverse process while specializing toward preferred and dispreferred directions.
Specifically, we fine-tune a winning model $\phi^{w}$ on the preferred samples $\{(\x_0^w,\vc)\}$ and a losing model $\phi^{l}$ on the dispreferred samples $\{(\x_0^l,\vc)\}$ using a standard diffusion training loss.
The learned policy pair provides stepwise contrastive score targets that incorporate the actual backward denoising transitions, given by $\noise_{\phi^{w}}(\x_t,t,\vc)$ and $\noise_{\phi^{l}}(\x_t,t,\vc)$, thereby avoiding forward-process-based approximations used in diffusion DPO--style methods.
Accordingly, we set the supervision targets in \cref{eq:DDSPO} as
\[
\noise_\star^{w} \;\leftarrow\; \noise_{\phi^{w}}(\x_t^w,t,\vc),
\qquad
\noise_\star^{l} \;\leftarrow\; \noise_{\phi^{l}}(\x_t^l,t,\vc),
\]
where $\x_t^w\!\sim\!q(\x_t\mid \x_0^w)$ and $\x_t^l\!\sim\!q(\x_t\mid \x_0^l)$.

\noindent \textbf{Training-Free Contrastive Policy Pair (TF-CPP)} \ \
Without preference-labeled data or additional training, we instantiate a contrastive policy pair by providing perturbed condition to a pretrained reference model to induce a losing direction; the resulting contrastive signals remain grounded in the reference model's reverse denoising process.
Specifically, we treat the denoising score predicted under the original prompt $\vc$ as the preferred signal, and construct a dispreferred counterpart by generating a corrupted prompt $\vc^{-}$ and computing the corresponding denoising score. 
Note that the type of perturbation applied to the prompt may vary depending on the specific application of preference optimization (\eg, text-to-image alignment or aesthetic quality).
In this setup, the noise $\noise_{\text{ref}}(\x_t^w, t, \vc)$ guided by $\vc$ is considered positively aligned with the intended semantics, while the noise $\noise_{\text{ref}}(\x_t^l, t, \vc^{-})$ from $\vc^-$ is dispreferred, as it reflects guidance from an incomplete or misleading prompt.

By plugging these noises as supervision targets into the DDSPO objective in \cref{eq:DDSPO}, we obtain the practical objective
\begin{align}
\mathcal{L}&(\theta) = 
- \mathbb{E}_{\substack{
(\x_0^w,\x_0^l) \sim \mathcal{D},\,
(\vc,\vc^{-})\sim \mathcal{D}(c), t \sim \mathcal{U}(0, T),\, \\\x_t^w 
\sim q(\x_t^w \mid \x_0^w), \, 
\x_t^l 
\sim q(\x_t^l \mid \x_0^l)
}}
\log \sigma \Bigr(- \beta \cdot [ \nonumber \\ &
\quad \| \noise_{\text{ref}}(\x_t^w, t, \vc) - \noise_\theta(\x_t^w, t, \vc) \|_2^2 \nonumber
- \| \noise_{\text{ref}}(\x_t^w, t, \vc) - \noise_{\text{ref}}(\x_t^w, t, \vc) \|_2^2 \nonumber \\ &- \big( 
\| \noise_{\text{ref}}(\x_t^l, t, \vc^{-}) - \noise_\theta(\x_t^l, t, \vc) \|_2^2 
- \| \noise_{\text{ref}}(\x_t^l, t, \vc^{-}) - \noise_{\text{ref}}(\x_t^l, t, \vc) \|_2^2 
\big)
] \Bigr)
\label{eq:practical_ddspo}
\end{align}
Here, the term \(\| \noise_{\text{ref}}(\x_t^w, t, \vc) - \noise_\theta(\x_t^w, t, \vc) \|_2^2 - \| \noise_{\text{ref}}(\x_t^w, t, \vc) - \noise_{\text{ref}}(\x_t^w, t, \vc) \|_2^2\) simplifies to a standard distillation loss, as the second term vanishes.
This encourages the student model to closely follow the reference without degrading its original capabilities.
In contrast, the term \(\| \noise_{\text{ref}}(\x_t^l, t, \vc^{-}) - \noise_\theta(\x_t^l, t, \vc) \|_2^2 - \| \noise_{\text{ref}}(\x_t^l, t, \vc^{-}) - \noise_{\text{ref}}(\x_t^l, t, \vc) \|_2^2\) penalizes the model when its prediction moves closer to the degraded direction \(\noise_{\text{ref}}(\x_t^l, t, \vc^{-})\) than the reference model's own output does. 
Together, these two terms encourage the model to retain useful denoising capabilities while avoiding alignment with poor or corrupted guidance signals.
Note that while constructing degraded prompts has been used in other domains to construct synthetic preference data, DDSPO is the first to leverage it directly as a contrastive policy pair, enabling stepwise supervision at every denoising step.

\noindent \textbf{Training Pipeline} \ \ 
DDSPO training depends on how the contrastive policy pair is instantiated. We describe the pipelines for DD-CPP and TF-CPP below.

\emph{DD-CPP.}
Starting from a preference-labeled dataset $\mathcal{D}=\{(\x_0^w,\x_0^l,\vc)\}$, we first obtain a contrastive policy pair by fine-tuning two models from the same pretrained diffusion initialization: a winning model $\phi^{w}$ on $\{(\x_0^w,\vc)\}$ and a losing model $\phi^{l}$ on $\{(\x_0^l,\vc)\}$ using a standard diffusion training loss.
We then perform DDSPO training by sampling $(\x_0^w,\x_0^l,\vc)\!\sim\!\mathcal{D}$, drawing $t\!\sim\!\mathcal{U}\{1,\dots,T\}$, and obtaining $\x_t^w\!\sim\!q(\x_t\mid \x_0^w)$ and $\x_t^l\!\sim\!q(\x_t\mid \x_0^l)$ via the forward process.
The stepwise supervision targets are computed from the learned policy pair as $\noise_\star^{w}\!\leftarrow\!\noise_{\phi^{w}}(\x_t^w,t,\vc)$ and $\noise_\star^{l}\!\leftarrow\!\noise_{\phi^{l}}(\x_t^l,t,\vc)$, and we optimize \cref{eq:DDSPO}.

\emph{TF-CPP.}
We construct condition pairs $(\vc,\vc^-)$ using task-specific condition perturbations (e.g., prompt perturbation for text-to-image alignment or aesthetics), and build a training set $\mathcal{D}=\{(\x_0^w,\x_0^l,\vc,\vc^-)\}$ by generating $(\x_0^w,\x_0^l)$ from a pretrained reference model under $\vc$ and $\vc^-$.
During training, we sample $(\x_0^w,\x_0^l,\vc,\vc^-)\!\sim\!\mathcal{D}$, draw $t\!\sim\!\mathcal{U}\{1,\dots,T\}$, obtain $\x_t^w\!\sim\!q(\x_t\mid \x_0^w)$ and $\x_t^l\!\sim\!q(\x_t\mid \x_0^l)$, and compute the targets $\noise_{\text{ref}}(\x_t^w,t,\vc)$ and $\noise_{\text{ref}}(\x_t^l,t,\vc^-)$ from the reference model.
We then optimize \cref{eq:practical_ddspo}.
We also consider an efficient TF-CPP variant that avoids generating dispreferred images $\x_0^l$ under $\vc^{-}$ by reusing a randomly sampled positive image to form $\x_t^l$ while still estimating the losing direction with $\vc^{-}$.
This is motivated by the observation~\cite{kim2025randomconditioningdistillationdataefficient} that, especially at larger timesteps, $\x_t$ is weakly dependent on $\x_0$ and the reverse denoising update is largely driven by the conditioning signal, so $\vc^{-}$ can still provide a meaningful negative score target even under prompt--image mismatch.
We report empirical results in \cref{sec:exp:perturbed}, with additional details provided in the 	\cref{sec:supp_perturbation}.

\subsection{Relationship to Previous Work}
\label{sec:method_prev_work}
\noindent \textbf{Diffusion DPO as a Special Case of DDSPO} \ \
While DDSPO uses 
\( p_\star^\ast(\x_{t-1}^\ast \mid \x_t^\ast, \vc) \) 
as the contrastive policy pair, 
Diffusion DPO is a special case of DDSPO 
that defines $p_\star^\ast(\x_{t-1}^\ast \mid \x_t^\ast, \vc)  
\;\triangleq\;
q(\x_{t-1}^\ast \mid \x_t^\ast, \x_0^\ast)$.
Under this definition, the stepwise supervision targets are induced from forward-process $q(\x_{t-1}^\ast \mid \x_t^\ast, \x_0^\ast)$ and thus ultimately depend on the terminal samples \( (\x_0^w,\x_0^l) \), rather than the model’s actual backward denoising transitions at inference time.
In contrast, DDSPO defines preference directly over local backward transitions \(\x_t\!\to\!\x_{t-1}\) under the winning and losing denoising policies \(p_\star(\x_{t-1}\mid \x_t,\vc)\),
yielding dense, transition-level supervision.

\noindent \textbf{Relationship to DSPO} \ \ 
DSPO~\cite{zhu2025dspo} reinterprets preference alignment through a 
score-matching formulation by constructing a preference-adjusted target score—
formed by combining the data score with a preference-derived score term—and 
fine-tuning the pretrained diffusion model to match this target.
Concretely, DSPO defines a score-matching style objective of the form
\begin{multline}
\label{eq:dspo_loss}
\mathcal{L}_{\text{DSPO}}(\theta)
= \mathbb{E}_{(\x_0^w,\x_0^l), \, \vc, \, t}A(t)
\Big\|
(\noise_\theta(\x_t^w,t,\vc) - \noise^w)
\\
\quad
- \beta
\Big(
1 - \sigma\big(
r_\theta(\x_t^w,t,\vc) - r_\theta(\x_t^l,t,\vc)
\big)
\Big)
\cdot (\noise_\theta(\x_t^w,t,\vc)
- \noise_\text{ref}(\x_t^w,t,\vc))
\Big\|_2^2.
\end{multline}
where $(\x_0^w,\x_0^l) \sim \mathcal{D},\, \vc \sim \mathcal{D}(c) ,\,
t \sim \mathcal{U}(0,T),\,
\x_t^w \sim q(\x_t^w \mid \x_0^w),\,
\x_t^l \sim q(\x_t^l \mid \x_0^l)$.
Here, $A(t)$ is a timestep-dependent weighting function determined by the diffusion noise schedule, 
and $r_\theta(\x_t,t,\vc)$ is defined as
\[
r_\theta(\x_t,t,\vc)
= 
\|\noise - \noise_\theta(\x_t, t, \vc)\|_2^2
- \|\noise - \noise_{\text{ref}}(\x_t, t, \vc)\|_2^2,
\]
where denoising score target $\noise \sim \mathcal{N}(0, I)$ is derived by approximating 
$p_\theta(\x_{t-1}\!\mid\!\x_t,\vc)$ 
with the forward process $ q(\x_{t-1} \mid \x_t, \x_0)$, 
so that the preference signal at timestep $t$ is still induced from preference-labeled final samples via a forward-process approximation.

In contrast, DDSPO removes reliance on the forward process by deriving preference supervision directly 
from the contrastive denoising policies—$p_\star^w(\x_{t-1}^w \!\mid\! \x_t^w, \vc)$ and 
$p_\star^l(\x_{t-1}^l \!\mid\! \x_t^l, \vc^-)$—which provide per-step, policy-grounded preference signals.
We can incorporate our per-step preference supervision into the DSPO by redefining
\[
r_\theta(\x_t,t,\vc)
= \|\noise_\star - \noise_\theta(\x_t, t, \vc)\|_2^2
- \|\noise_\star - \noise_{\text{ref}}(\x_t, t, \vc)\|_2^2,
\]
where $\noise_\star^w$ and $\noise_\star^l$ denote denoising score targets derived from the contrastive policy pair—the winning and losing denoising policies $p_\star^w(\x_{t-1}^w \mid \x_t^w, \vc)$ and $p_\star^l(\x_{t-1}^l \mid \x_t^l, \vc)$, respectively, as defined in DDSPO.
In practice, these targets can be obtained from the reference model as 
$\noise_{\text{ref}}(\x_t^w, t, \vc)$ and $\noise_{\text{ref}}(\x_t^l, t, \vc^-)$.
This reformulation extends DSPO to operate in a stepwise manner, providing transition-level supervision grounded in the model’s actual backward denoising trajectory rather than forward-process surrogates. The detailed derivation and empirical results are provided in \cref{sec:supp_derivation} and \cref{sec:exp:align}, respectively.

\noindent \textbf{Relationship to SPO} \ \
Step-by-step Preference Optimization (SPO)~\cite{liang2025aestheticposttrainingdiffusionmodels} also provides stepwise supervision, but it does so by sampling multiple candidates from a shared noisy latent $\x_t$ and selecting a win--lose pair $(\x_{t-1}^w,\x_{t-1}^l)$ using a step-aware reward model trained to assess visual quality. 
Since this reward model is trained using preference signals constructed through the forward process, it may provide a imperfect reward signal for the model’s actual backward denoising transitions.
Consequently, SPO's step-aware reward tends to prioritize subtle, fine-grained visual details (\eg, texture, lighting, color tone) over high-level semantic or layout features, which leads to limited improvements in text--image alignment, as reported in the original paper~\cite{liang2025aestheticposttrainingdiffusionmodels} \dohyun{and further evaluated under our setup in the \cref{sec:supp_additional_comparisons}}.
Moreover, SPO relies on stochastic multi-candidate sampling for per-step selection, which limits its applicability to deterministic flow-matching models~\cite{lipman2023flow, xie2024sana}.
In contrast, DDSPO provides stepwise supervision without relying on a step-aware reward model or stochastic multi-candidate sampling (and per-step selection) as in SPO.
Instead, DDSPO derives contrastive score targets aligned with the model's actual denoising trajectory by defining stepwise supervision directly from a contrastive policy pair.
This formulation is effective for both aesthetic enhancement and text--to-image alignment, and can be applied to both diffusion and flow-based generative models.
We empirically validate these properties across diverse architectures and tasks in \cref{sec:experiments}.
\section{Experiments}
\label{sec:experiments}
We assess the effectiveness of DDSPO on a range of conditional generation tasks, focusing on its ability to improve alignment and perceptual quality. \dohyun{It is important to note that, for text-to-image alignment, there is no publicly available preference-labeled dataset; accordingly, we instantiate the contrastive policy pair using TF-CPP.}
In this section, we begin with a controlled 2D toy experiment (\cref{sec:exp:toy}) to validate DDSPO’s core mechanism under minimal conditions. We then apply DDSPO to two practical text-to-image tasks: improving prompt-image alignment (\cref{sec:exp:align}) and enhancing aesthetic quality (\cref{sec:exp:aesthetic}). 
Finally, we explore different prompt degradation strategies for TF-CPP, together with an efficient variant (\cref{sec:exp:perturbed}).
\dohyun{Unless otherwise specified, we train for 100 iterations with an effective batch size of 2,048; further implementation details are provided in the \cref{sec:supp_impl}.}
\previous{Implementation details are provided in the \cref{sec:supp_impl}.}

\begin{figure*}[t]
    \centering
    \begin{subfigure}[b]{0.152\textwidth}
        \includegraphics[width=\textwidth]{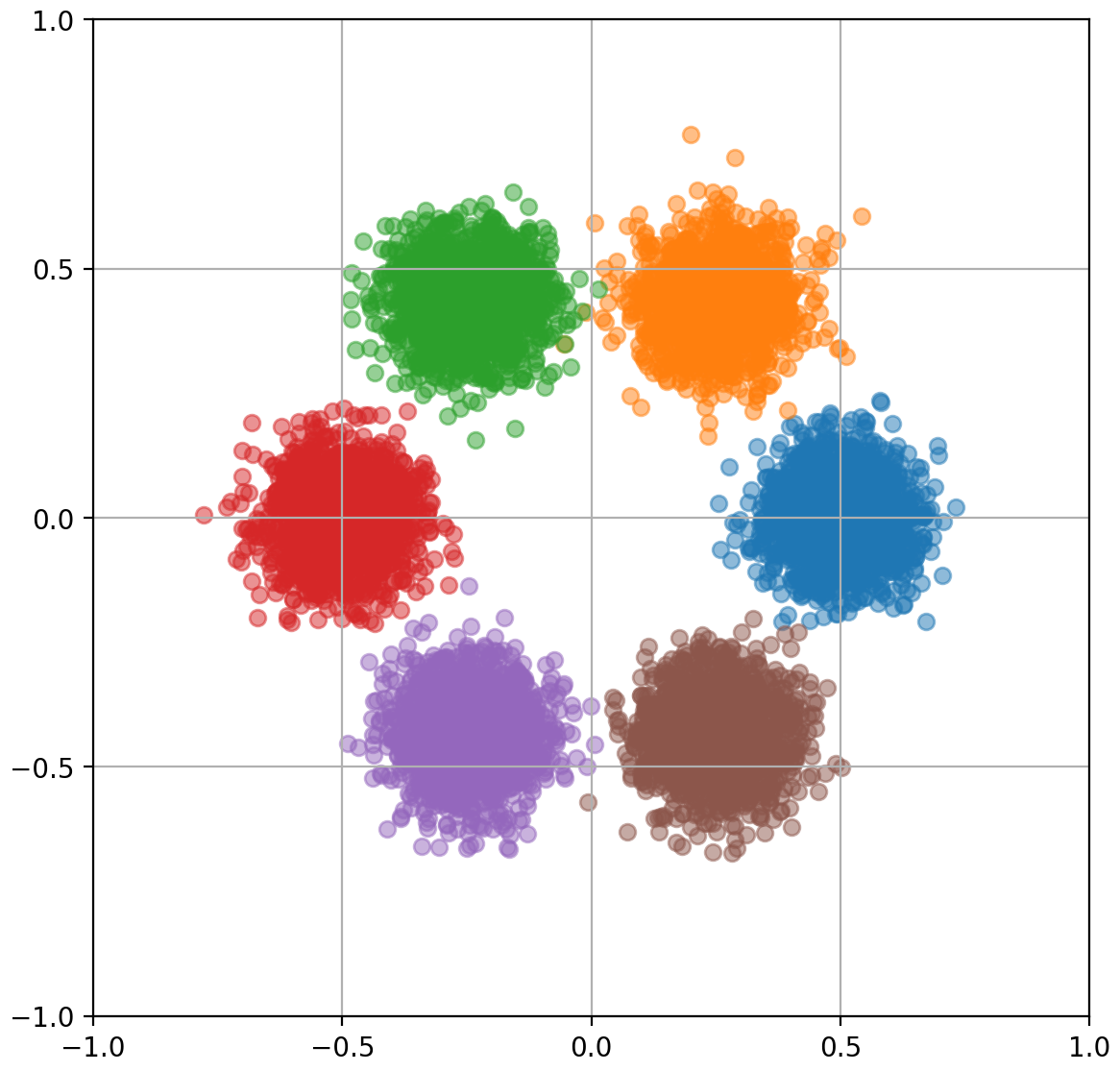}
        \caption{GT}
        \label{fig:toy_results:gt}
    \end{subfigure}
    \begin{subfigure}[b]{0.152\textwidth}
        \includegraphics[width=\textwidth]{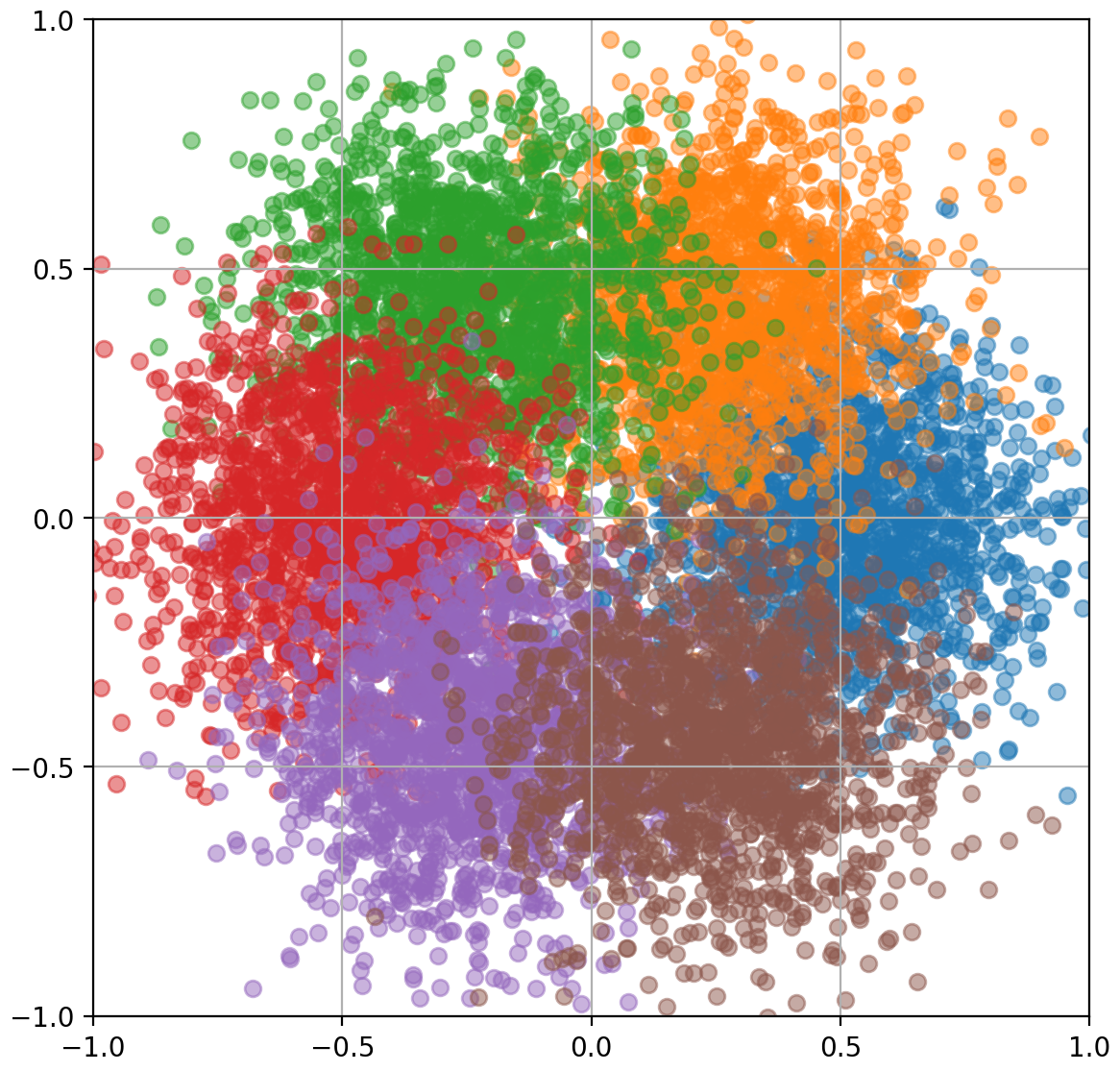}
        \caption{Train Data}
        \label{fig:toy_results:train}
    \end{subfigure}
    \begin{subfigure}[b]{0.152\textwidth}
        \includegraphics[width=\textwidth]{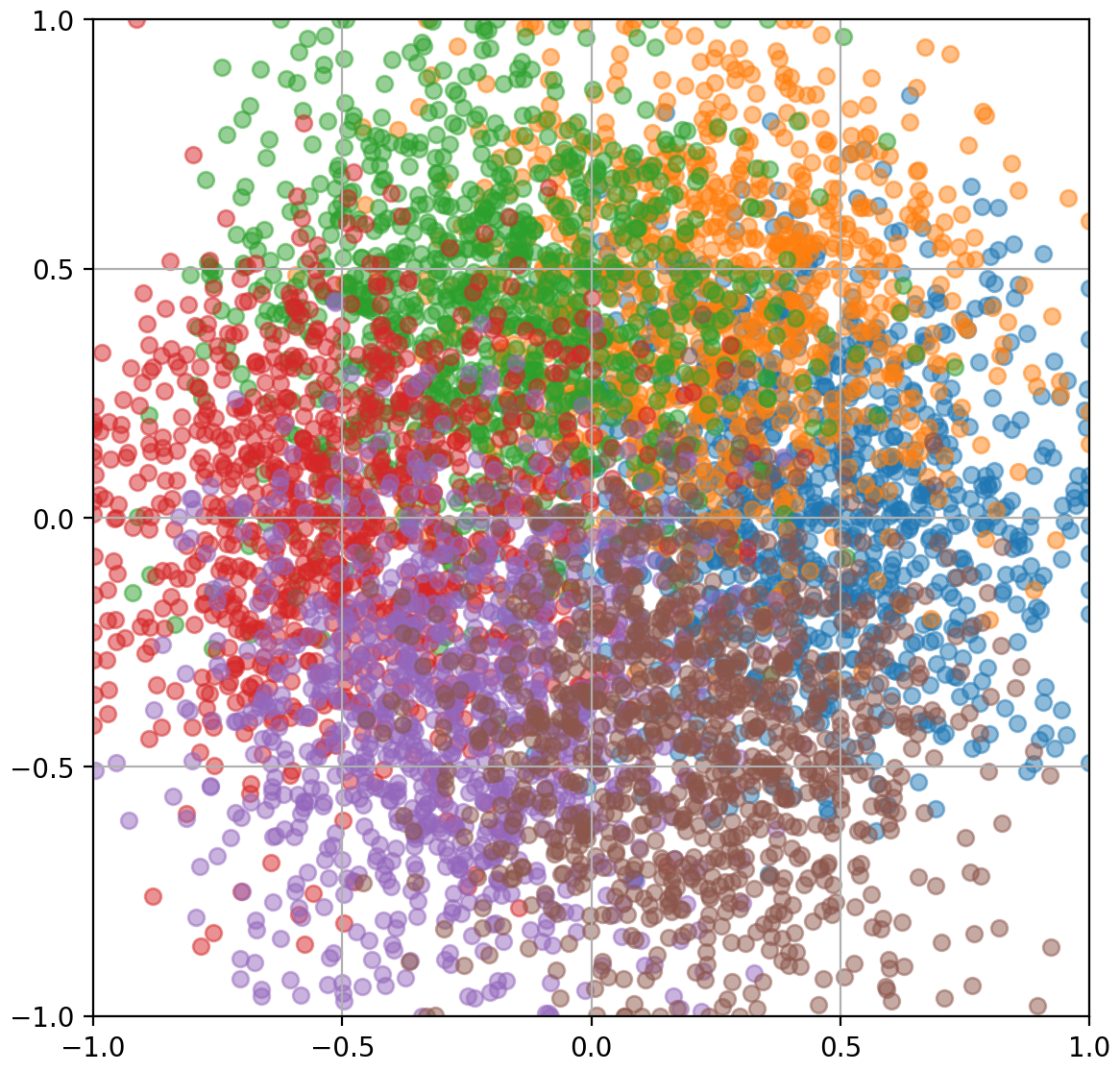}
        \caption{Ref Model}
        \label{fig:toy_results:ref}
    \end{subfigure}
    \begin{subfigure}[b]{0.152\textwidth}
        \includegraphics[width=\textwidth]{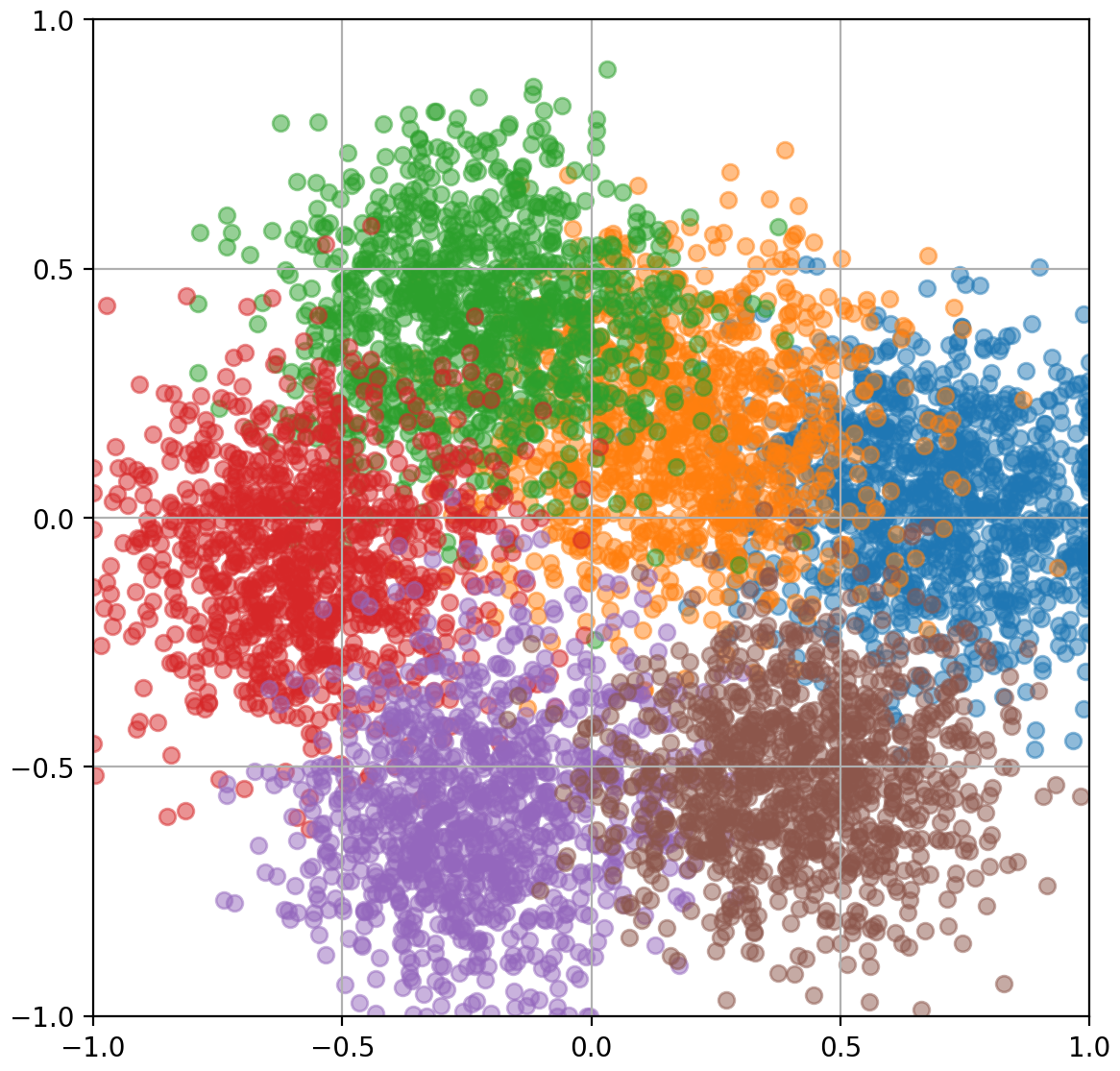}
        \caption{D-DPO}
        \label{fig:toy_results:dpo}
    \end{subfigure}
    \begin{subfigure}[b]{0.152\textwidth}
        \includegraphics[width=\textwidth]{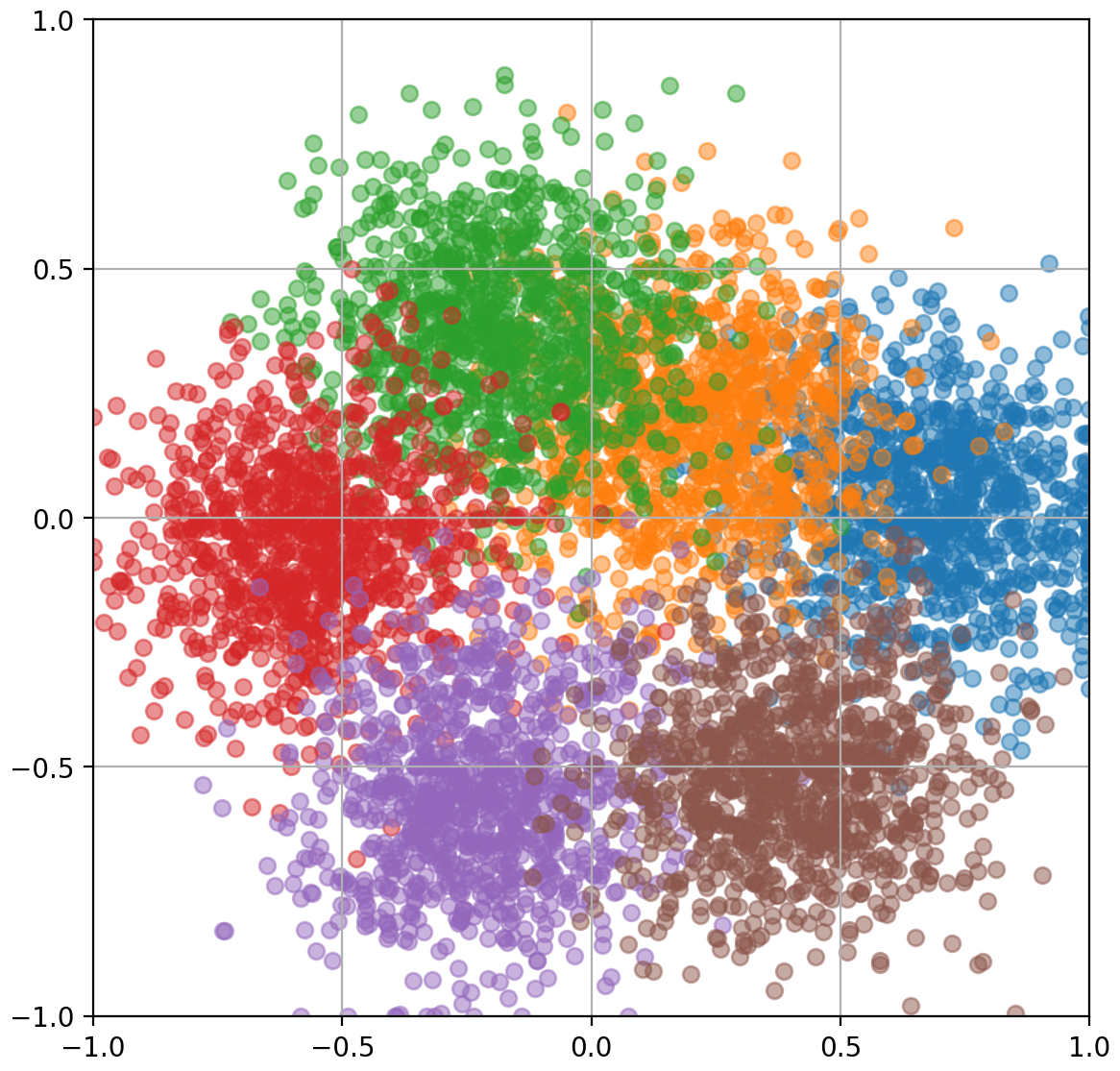}
        \caption{DSPO}
        \label{fig:toy_results:dspo}
    \end{subfigure}
    \begin{subfigure}[b]{0.152\textwidth}
        \includegraphics[width=\textwidth]{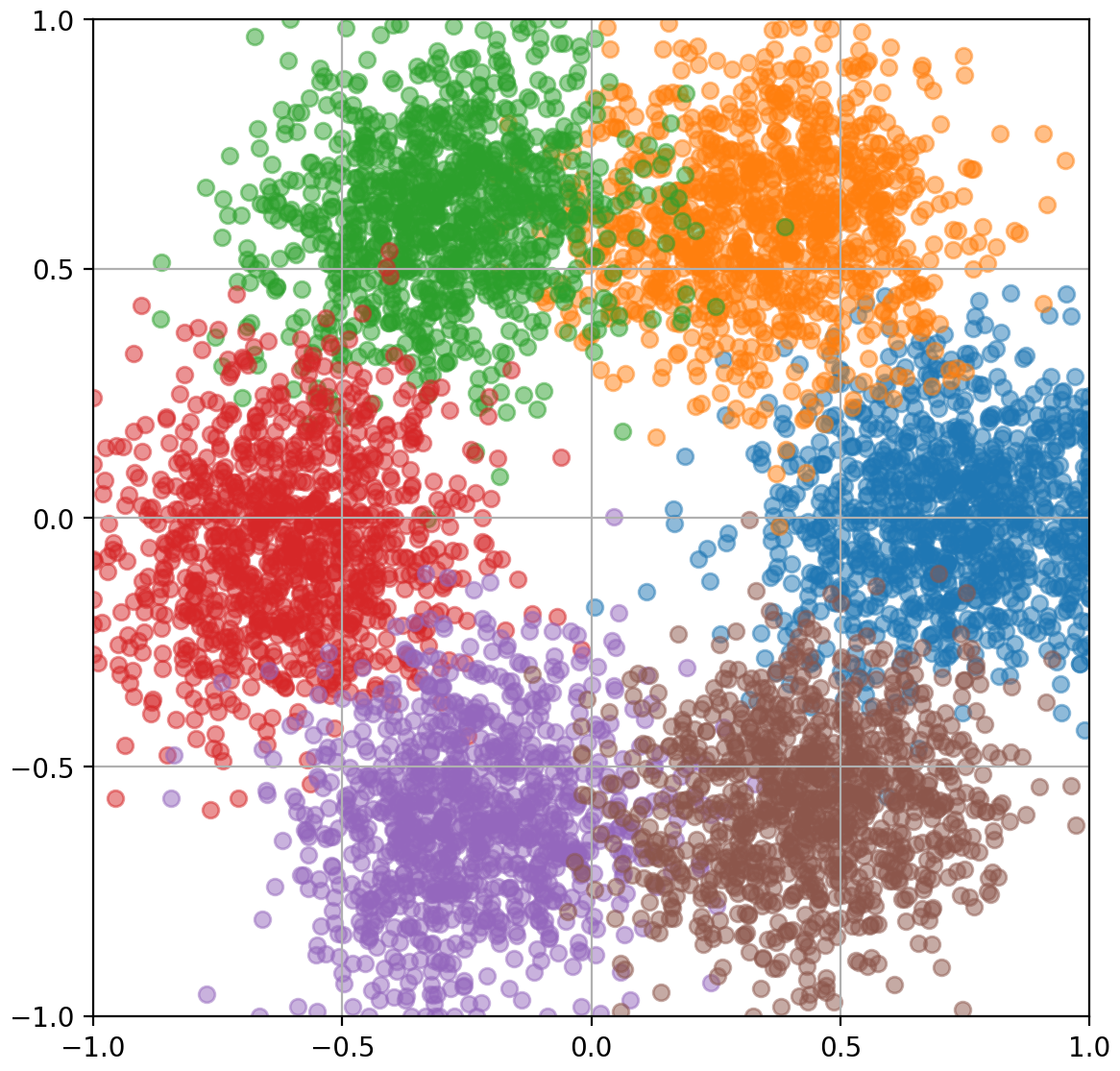}
        \caption{DDSPO}
        \label{fig:toy_results:DDSPO}
    \end{subfigure}
    \vspace{-4pt}
    \caption{\textbf{Toy Experiments Comparison of Diffusion DPO~(D-DPO), DSPO and DDSPO.}
    (a) and (b) show samples from the ground-truth distribution and its noisy variant used for training. 
    (c) is generated by the reference diffusion model trained on (b). 
    (d), (e) and (f) are distributions learned by the models finetuned with Diffusion DPO, DSPO  and DDSPO, respectively.
    }
    \label{fig:toy_results}
  \vspace{-8pt}
\end{figure*}
\subsection{Preliminary Experiments}
\label{sec:exp:toy}

We first conduct toy experiments to evaluate the effectiveness of DDSPO in a controlled 2D setting. 
Specifically, we simulate a simplified conditional generation task where each condition corresponds to a distinct mode of a multi-modal Gaussian distribution, as illustrated in \cref{fig:toy_results:gt}.
A reference model is trained on a noisy dataset shown in \cref{fig:toy_results:train}, mimicking real-world scenarios with imperfect supervision. 
The learned distribution from this reference model, shown in \cref{fig:toy_results:ref}, closely resembles the noisy training distribution.
To construct preference data, we sample $N = 2$ preference pairs per class, each consisting of a preferred sample $\x_0^w$ from the target class $\vc$ and a dispreferred sample $\x_0^l$ from a neighboring class $\vc^-$, simulating our perturbed prompting strategy.
This dataset is then used to finetune the model using Diffusion DPO (D-DPO)~\cite{Wallace_2024_CVPR}, DSPO~\cite{zhu2025dspo}, and DDSPO.
Among them, D-DPO and DSPO serve as baselines that obtain supervision only from preference-labeled final samples, 
whereas DDSPO leverages intermediate score preferences at each timestep $t$.

As shown in \cref{fig:toy_results:dpo,fig:toy_results:dspo}, the baseline models finetuned with D-DPO and DSPO often fail to maintain clear separation between modes, resulting in overlapping or distorted output distributions.
This issue becomes more pronounced 
when the fine-tuning dataset is smaller (see \cref{sec:supp_extended} for extended results).
The degradation arises because these baselines rely on denoising targets derived from potentially misaligned final samples, which can provide misleading learning signals as D-DPO always provides noise towards this particular final sample, as discussed in \cref{sec:method:DDSPO}.
In contrast, DDSPO enables the model to learn well-separated, condition-specific outputs, as shown in \cref{fig:toy_results:DDSPO}.
Rather than relying on a final samples (which can induce global, coarse signals), DDSPO directly models preference at each $t$ by contrasting the winning and losing denoising policies $p_\star^w$ and $p_\star^l$ over the local transition $(\x_t \!\to\! \x_{t-1})$.
This leads to more robust learning signals across the trajectory of denoising.
Note that in the \cref{fig:toy_results:DDSPO}, slight shifts of the centers away from zero can occur because DDSPO separates distributions across conditions by following the ``winning'' direction while pushing away from the ``losing'' (neighboring) direction in score space, yielding clearer boundaries between adjacent conditional distributions. In more complex, real-world tasks such as image generation, the condition space (\eg, text space) is far more densely packed, providing supervision that contrasts across a wider range of diverse directions.

\subsection{Improving Text-to-Image Alignment}
\label{sec:exp:align}

\noindent \textbf{Target Task} \ \ 
We evaluate DDSPO for improving text-to-image alignment, enabling more faithful generation with respect to the input prompt.

\noindent \textbf{Experimental Setup} \ \ 
As no publicly available preference-labeled dataset exists for text-to-image alignment, we instantiate the contrastive policy pair using TF-CPP.
For a fair comparison, we train D-DPO and DSPO on the same TF-CPP-constructed preference pairs $(\x_0^w,\x_0^l)$.
We sample 200K prompts from DiffusionDB~\cite{diffusionDB} as the original conditions $\vc$; for each $\vc$, we generate a perturbed caption $\vc^-$ by randomly removing 40\% to 70\% of the tokens, thereby reducing semantic specificity and content richness.
To assess text-to-image alignment, we evaluate each model on the GenEval~\cite{ghosh2023geneval} and T2I-CompBench~\cite{huang2023t2i} benchmarks, which are designed to measure compositional and semantic consistency between text prompts and generated images. Additionally, we compute the Fréchet Inception Distance (FID)~\cite{FID} and Inception Score (IS)~\cite{IS} on 30K images generated from MS-COCO~\cite{mscoco} validation prompts to assess output quality and diversity.

\begin{table}[t]
\centering
\caption{
\textbf{Comparison of Preference Optimization Methods.} 
We evaluate D-DPO, DSPO, DSPO+CPP (our stepwise reformulation of DSPO with contrastive policy pair (CPP) described in 
\cref{sec:method_prev_work}), and DDSPO across GenEval, 
T2I-CompBench, FID, and IS, using identical TF-CPP-constructed preference data.
}
\label{tab:align_comp_dpo}
\vspace{-0.2cm}
\scalebox{0.85}{
\begin{tabular}{lccccc}
    \hline
    \hline
    Model &  GenEval$\uparrow$ & CompBench$\uparrow$ & FID$\downarrow$ & IS$\uparrow$ \\
    \hline
    \rowcolor{gray!25}
    SD-1.4 & .4245 & .3150 & 13.05 &36.76 \\
    ~+D-DPO & .4841 & .3723 & 18.02 &36.64 \\
    ~+DSPO & .4841 & .3793 & 18.28 &35.87 \\
    ~+DSPO+CPP (Ours) & .4978 & .3854 & 17.12 & 37.47 \\
    ~+DDSPO (Ours) & .5045 & .3823 & 16.39 &38.10 \\
    \hline
    \hline
\end{tabular}
}
\vspace{-0.2cm}
\end{table}

\begin{figure*}[t]
    \centering
    \includegraphics[width=1\linewidth]{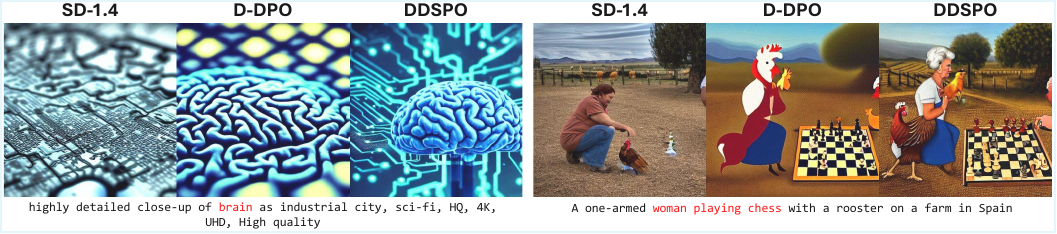}
    \vspace{-0.7cm}
    \caption{\textbf{Qualitative Comparison between SD-1.4, D-DPO and DDSPO.} D-DPO and DDSPO are fine-tuned using the same TF-CPP-constructed preference data as in \cref{tab:align_comp_dpo}. Images are generated from the same prompts and random seeds.}
    \label{fig:qualitative_sd14_dpo}
\end{figure*}

\noindent \textbf{Comparisons of Preference Optimization Methods} \ \ 
We finetune Stable Diffusion v1.4 (SD-1.4)~\cite{sdm_v1.4} using D-DPO~\cite{Wallace_2024_CVPR}, DSPO~\cite{zhu2025dspo}, DSPO+CPP (our stepwise reformulation of DSPO with contrastive policy pair (CPP) described in 
\cref{sec:method_prev_work}), and DDSPO, and compare their performance in \cref{tab:align_comp_dpo}.
\dohyun{Both CPP-based variants outperform their corresponding forward-process-based baselines: DDSPO achieves the best GenEval score, while DSPO+CPP obtains the best T2I-CompBench score.}
\previous{While all methods exhibit performance gains, the proposed DDSPO achieves the most substantial improvements in alignment metrics (GenEval and T2I-CompBench).}
Furthermore, it attains higher IS, as the strengthened alignment encourages the model to generate more distinct and well-separated representations in the feature space.
\dohyun{We also observe increased FID across all preference-optimization methods, partly attributable to the distribution gap between DiffusionDB-based training images and the MS-COCO evaluation set. DDSPO nevertheless shows the smallest increase while retaining strong alignment gains; further analysis is provided in the \cref{sec:supp_extended}.}
\previous{We also observe that fine-tuning with preference optimization generally introduces domain shifts from the MS-COCO image distribution; however, DDSPO remains closer to the original distribution while achieving the largest alignment improvements.}
\dohyun{Finally, replacing forward-posterior targets with CPP-derived stepwise targets also improves DSPO, showing that the benefit of CPP-based supervision extends beyond the DPO objective.}
\previous{Finally, incorporating our score-based preferences into DSPO also enhances performance, highlighting the advantage of score-level supervision even for preference optimization in the score-matching framework.}
Qualitative results comparing the baseline SD-1.4 model, D-DPO, and the DDSPO-finetuned model can be found in \cref{fig:qualitative_sd14_dpo}.

\begin{table*}[t]
\centering
\caption{\textbf{Text-to-Image Alignment Evaluation.}
(a) We evaluate DDSPO across diverse backbones (SD-1.4~\cite{sdm_v1.4}, SDXL~\cite{SDXL}, SANA~\cite{xie2024sana}, SD3-M~\cite{sd3}).
(b) Following CaPO's evaluation protocol, we compare SDXL with Itercomp, CaPO, and DDSPO variants.
}
\vspace{-0.5cm}
\label{tab:align_comp}
\begin{minipage}[t]{0.38\textwidth}
    \centering
    \subcaption{DDSPO on diverse backbones}
    \label{tab:align_comp_models}
    \scalebox{0.85}{
    \begin{tabular}{lcc}
        \hline
        \hline
        Model & GenEval$\uparrow$ & Compbench$\uparrow$ \\
        \hline
        \rowcolor{gray!25} SD-1.4 & .4245 & .3150 \\
        ~+DDSPO                    & .5045 & .3723 \\
        \hline
        \rowcolor{gray!25} SDXL  & .5229 & .4034 \\
        ~+DDSPO                     & .6049 & .4857 \\
        \hline
        \rowcolor{gray!25} SANA  & .6812 & .4846 \\
        ~+DDSPO                     & .7266 & .5255 \\
        \hline
        \rowcolor{gray!25} SD3-M  & .7123 & .5155 \\
        ~+DDSPO                     & .7384 & .5740 \\
        \hline
        \hline
    \end{tabular}
    }
\end{minipage}
\hfill
\begin{minipage}[t]{0.58\textwidth}
    \centering
    \subcaption{Comparisons to SOTA methods}
    \label{tab:align_comp_methods}
    \scalebox{0.85}{
    \begin{tabular}{lccc}
        \hline
        \hline
        Model & Human Annotation & GenEval$\uparrow$ & Compbench$\uparrow$ \\
        \hline
        \rowcolor{gray!25} SDXL & - & .5229 & .4185 \\
        ~+Itercomp    & \ding{51} & .6108 & .4644 \\
        ~+CaPO        & \ding{51} & .5900 & .4652 \\
        ~+DDSPO       & \ding{55} & .6049 & .5064 \\
        \hline
        \rowcolor{gray!25} SD3-M & - & .7123 & .5155 \\
        ~+CaPO        & \ding{51} & .7100 & .5238 \\
        ~+DDSPO       & \ding{55} & .7384 & .5740 \\
        \hline
        \hline
    \end{tabular}
    }
\end{minipage}
\end{table*}


\noindent \textbf{DDSPO with Various Model Architectures} \ \ 
We apply DDSPO to models with different architectures, including U-Net-based models (SD-1.4~\cite{sdm_v1.4} and SDXL~\cite{SDXL}) and DiT-based flow-matching models (SANA~\cite{xie2024sana} and SD3-M~\cite{sd3}), as shown in \cref{tab:align_comp_models}.
The results clearly demonstrate that DDSPO significantly improves text-to-image alignment across all models on both benchmarks, highlighting its effectiveness and broad applicability across diverse diffusion architectures.

\noindent \textbf{Comparisons to SOTA Methods} \ \ 
In \cref{tab:align_comp_methods}, we compare our method against current SOTA approaches, Itercomp~\cite{zhang2025itercomp} and CaPO~\cite{lee2025calibratedmultipreferenceoptimizationaligning}, for improving text-to-image alignment.
Notably, IterComp and CaPO rely on human-annotated datasets or reward models for training.
In contrast, DDSPO achieves comparable performance on GenEval and outperforms all methods on CompBench, \dohyun{without human-annotated preference data or preference-trained reward models} \previous{despite not using any human annotations}, demonstrating its effectiveness in a data efficient setup.

\subsection{Improving Aesthetic Quality}
\label{sec:exp:aesthetic}

\noindent \textbf{Target Task} \ \ 
In this section, we apply DDSPO to improve the aesthetic quality of images generated by diffusion models, promoting visually appealing and artistically coherent outputs.

\noindent \textbf{Experimental Setup} \ \ 
\dohyun{We train on SD-1.5~\cite{sdm_v1.5} and SDXL~\cite{SDXL}, and evaluate DDSPO under both DD-CPP (Pick-a-Pic~\cite{pickscore}) and TF-CPP, where TF-CPP constructs an aesthetically degraded counterpart $\vc^{-}$ using LLaMA3-8B~\cite{grattafiori2024llama3herdmodels}.}
The prompt template used for generation is provided in the 	\cref{sec:supp_perturbation}.
To evaluate aesthetic quality, we use the HPSv2~\cite{wu2023human} and PickScore~\cite{pickscore} metrics, each measured by a model trained to reflect human aesthetic preferences using Human Preference Dataset v2~\cite{wu2023human} and Pick-a-Pic~\cite{pickscore}, respectively.
For evaluation, we generate images by sampling from prompts in the HPSv2~\cite{wu2023human} and PartiPrompts~\cite{yu2022scaling} benchmarks, and assess them using the corresponding metrics.


\begin{table*}[t]
\centering
\caption{\textbf{Aesthetic Quality Comparison with SOTA Methods.}
\dohyun{We compare DDSPO with existing methods for aesthetic quality improvement, using DD-CPP with Pick-a-Pic and TF-CPP without human-annotated preference data.}
\previous{We compare our DDSPO with existing methods that rely on human-labeled supervision from Pick-a-Pic, whereas DDSPO is trained without human-labeled data.}
}
\vspace{-0.4cm}
\label{tab:aesthetic_sota}
\begin{minipage}[t]{0.49\textwidth}
    \centering
    \subcaption{Comparisons to SOTA methods with SD-1.5}
    \label{tab:aesthetic_sd15}
    \scalebox{0.85}{
    \begin{tabular}{lccc}
        \hline
        \hline
        Model & Dataset & HPSv2$\uparrow$ & PickScore$\uparrow$  \\
        \hline
        \rowcolor{gray!25}
        SD-1.5 & - & 26.95 & 21.14 \\
        ~+D-DPO &Pickapic& 27.25 & 21.34\\
        ~+D-KTO &Pickapic& 27.89 & 21.39  \\
        ~+SPO & Pickapic & 27.50 & 21.41 \\
        ~+DDSPO & Pickapic & 27.60 & 21.50 \\
        ~+DDSPO & None & 27.46 & 21.35 \\
        \hline
        \hline
    \end{tabular}
    }
\end{minipage}
\hfill
\begin{minipage}[t]{0.49\textwidth}
    \centering
    \subcaption{Comparisons to SOTA methods with SDXL}
    \label{tab:aesthetic_sdxl}
    \scalebox{0.85}{
    \begin{tabular}{lccc}
        \hline
        \hline
        Model & Dataset & HPSv2$\uparrow$ & PickScore$\uparrow$  \\
        \hline
        \rowcolor{gray!25}
        SDXL & - & 27.89 & 22.27\\
        ~+D-DPO & Pickapic & 28.55 & 22.61 \\
        ~+MAPO & Pickapic & 28.22 & 22.30 \\
        ~+SPO & Pickapic & 29.21 & 23.11 \\
        ~+DDSPO & Pickapic & 29.09 & 22.71 \\
        ~+DDSPO & None & 28.78 & 22.70 \\
        \hline
        \hline
    \end{tabular}
    }
\end{minipage}
\end{table*}

\noindent \textbf{Results} \ \
In \cref{tab:aesthetic_sota}, we compare DDSPO with SOTA methods for aesthetic quality improvement, including D-DPO~\cite{Wallace_2024_CVPR}, Diffusion KTO~(D-KTO)~\cite{diffkto}, SPO~\cite{liang2025aestheticposttrainingdiffusionmodels} and MAPO~\cite{hong2025marginaware}.
\dohyun{Under matched Pick-a-Pic supervision, DDSPO outperforms D-DPO on both backbones. TF-CPP also remains competitive with supervised methods without human-annotated preference data, although D-KTO and SPO perform better on some metrics.}
\previous{DDSPO achieves competitive or superior performance across 
both settings (DD-CPP and TF-CPP), consistently outperforming D-DPO under 
the same training data, demonstrating the effectiveness 
of stepwise contrastive supervision over 
forward process based approximations.}

\subsection{Contrastive Policy Pair Construction Strategies}
\label{sec:exp:perturbed}
\noindent \textbf{Effects of Different Prompt Degradation Strategies} \ \ 
\cref{tab:perturbation_strategies} presents an exploration of different prompt degradation strategies for constructing dispreferred scores across the two previously studied tasks. 
We evaluate three strategies: \textit{Rand-removal}, which drops 40\% to 70\% of tokens from the original prompt; 
and \textit{LLaMA}, which uses an LLM~\cite{grattafiori2024llama3herdmodels} to rewrite the prompt in a way that degrades task performance.
The LLaMA-based strategy includes two task-specific variants: LLaMA (TA) for text-to-image alignment and LLaMA (AQ) for aesthetic quality improvement.
When trained with DDSPO, we observe that Rand-removal yields the best performance on text-to-image alignment, while LLaMA (AQ) performs best for aesthetic quality improvement, as it is specifically tailored to that task.

\begin{table*}[t]
\centering
\caption{
\textbf{Effects of Various Prompt Degrading Strategies and Negative Image Generation.}
Ablation study comparing (a) different strategies for constructing dispreferred prompts to generate negative samples, and (b) standard DDSPO with variants that avoid explicit negative image~(NI) generation by reusing preferred samples with unpaired or perturbed prompts.
}
\vspace{-0.4cm}
\begin{minipage}[t]{0.55\textwidth}
    \centering
    \subcaption{Comparisons of different perturbation Strategies}
    \label{tab:perturbation_strategies}
    \scalebox{0.85}{
    \begin{tabular}{lcccc}
        \hline
        \hline
        Method & \multicolumn{2}{c}{T2I Alignment} &  \multicolumn{2}{c}{Aesthetic Quality} \\
        \cmidrule(lr){2-3} \cmidrule(lr){4-5}&GE$\uparrow$ & CB$\uparrow$ & HPSv2$\uparrow$ & PickScore$\uparrow$ \\
        \hline
        \rowcolor{gray!25}
        SD-1.4 & .4245 & .3150 & 26.88 & 21.11\\
        Rand-removal & .5045 & .3823 & 27.28 & 21.36 \\
        LLaMA (TA) & .4854 & .3807 & 27.25 & 21.36 \\
        LLaMA (AQ) & .4758& .3616 & 27.51 & 21.39 \\
        \hline
        \hline
    \end{tabular}
    }
\end{minipage}
\hfill
\begin{minipage}[t]{0.43\textwidth}
    \centering
    \subcaption{Comparison of DDSPO with and without explicit negative image generation}
    \label{tab:neg_modeling}
    \scalebox{0.85}{
    \begin{tabular}{lccc}
        \hline
        \hline
        Method & NI generation & GE$\uparrow$ & CB$\uparrow$ \\
        \hline
        \rowcolor{gray!25}
        SD-1.4 & - & .4245 & .3150 \\
        Rand-removal & \ding{51} & .5045 & .3823 \\
        Rand-positive & \ding{55} & .4866 & .3763 \\
        Not-paired & \ding{55} & .4891 & .3842 \\
        \hline
        \hline
    \end{tabular}
    }
\end{minipage}
\vspace{-0.2cm}
\label{tab:neg_modeling_combined}
\end{table*}
\noindent \textbf{Efficient DDSPO} \ \ 
We introduce a data-efficient variant of DDSPO in which only preferred images $\x_0^w$ are generated, without dispreferred ones $\x_0^l$, resulting in a dataset $\mathcal{D}={(\x_0^w, \vc)}$ instead of $(\x_0^w,\x_0^l,\vc,\vc^-)$.
To apply DDSPO with $\mathcal{D}$, however, we still require a dispreferred sample $\x_0^l$ to obtain its noisy version $\x_t^l$ through the forward diffusion process, which is necessary to compute the dispreferred score $\noise_\star^l$.
Following the technique described in \cref{sec:construct_score}, we randomly sample an image $\x_0^+$ from $\mathcal{D}$ that was generated by a prompt unrelated to $\vc$, and designate it as a pseudo-dispreferred image.
Using this setup, we evaluate two perturbation strategies: Random-positive, which pairs the original prompt with a randomly selected image $\x_0^+$; and Not-paired, which uses a perturbed version of the original prompt (via random token removal) that is not strictly paired with the noised image $\x_t^l$.
We compare these variants to the full setup that explicitly generates dispreferred images paired with perturbed prompts (Rand-Removal).
As shown in \cref{tab:neg_modeling}, both Random-positive and Not-paired perform comparably to Rand-removal, where explicitly generated dispreferred samples are available.
Interestingly, Not-paired outperforms Random-positive, despite using unpaired image–prompt combinations.
These findings demonstrate that the negative direction in DDSPO can be effectively modeled from unpaired prompt–image instances, underscoring the framework’s flexibility in preference-signal modeling via stepwise supervision.

\section{Conclusion}
\label{sec:conclusion}
We introduce Direct Diffusion Score Preference Optimization (DDSPO), which derives stepwise supervision targets from a contrastive policy pair, grounding supervision in the model’s backward denoising transitions rather than the forward-process posterior \(q(\x_{t-1}\mid \x_t,\x_0)\).
We instantiate the policy pair via either DD-CPP (preference-labeled data) or TF-CPP (a pretrained reference model with an original condition and a semantically degraded counterpart).
Across text--image alignment and aesthetic quality tasks, DDSPO consistently improves its base models, demonstrating the effectiveness of stepwise supervision over backward denoising transitions.

\section*{Acknowledgements}
This research was supported by the IITP grants (IITP-2026-RS-2024-00436857, IITP-2026-RS-2020-II201819, IITP-2026-RS-2025-02304828, and RS-2026-25507282), the NRF grant (RS-2025-23523979), and the KOCCA grants (RS-2024-00345025 and RS-2026-25506607), funded by the Korea government (MSIT and MCST).



%
%
\bibliographystyle{splncs04}
\bibliography{main}

\onecolumn
\appendix
\begin{center}
  \vspace{0.25in}

  {\LARGE \bfseries Direct Diffusion Score Preference Optimization via Stepwise Contrastive Policy-Pair Supervision \par}

  \vspace*{24pt}
  {\Large\itshape supplementary material}

  \vspace{0.29in}
  \vspace{0.09in}
\end{center}

\renewcommand{\thetable}{\Alph{table}}
\renewcommand{\thefigure}{\Alph{figure}}
\setcounter{table}{0}
\setcounter{figure}{0}
\section{Derivation}
\label{sec:supp_derivation}

\noindent\textbf{Diffusion Model}
Denoising diffusion probabilistic models~\cite{ddpm, sohl2015deep, song2021scorebased} define a generative process that reverses a fixed forward noising process applied to data drawn from $q(\x_0)$. This reverse process is modeled as a discrete-time Markov chain over latent variables $\x_{0:T}$, with the joint distribution defined as $p_\theta(\x_{0:T}) = p(\x_T) \prod_{t=1}^T p_\theta(\x_{t-1} \mid \x_t)$.
Each transition $p_\theta(\x_{t-1} \mid \x_t)$ is modeled as a Gaussian distribution:
\begin{equation}\label{eq:reverse_likelihood}
    p_\theta(\x_{t-1}|\x_t)=\calN(\x_{t-1};\mu_\theta(\x_t), \sigma_{t|t-1}^2\frac{\sigma_{t-1}^2}{\sigma_t^2}\I).
\end{equation}
During training, the model is optimized to approximate the true reverse process by minimizing a variational bound on the negative log-likelihood. In practice, this reduces to predicting the noise $\boldsymbol{\epsilon}$ added at each timestep, leading to a simplified objective of the form:
\begin{equation}
    \mathcal{L}_{\text{MSE}} = \mathbb{E}_{t, \x_0, \boldsymbol{\epsilon} \sim \mathcal{N}(0, \I)} \left[ \left\| \boldsymbol{\epsilon} - \boldsymbol{\epsilon}_\theta(\x_t, t) \right\|_2^2 \right],
\end{equation}
where $\x_t = \sqrt{\alpha_t} \x_0 + \sqrt{1 - \alpha_t} \boldsymbol{\epsilon}$ is the noised sample at timestep $t$, and $\boldsymbol{\epsilon}_\theta$ is the model's prediction of the noise.

\noindent\textbf{Direct Preference Optimization} \ \ 
In \cref{equ:BT_distribution}, reward model \( r(\vc,\xw_0) \) can be parameterized by a nerual net work $\phi$ and estimated via maximum likelihood training for binary classification: 

\begin{equation}
    L_\text{BT}(\phi) = - 
    \bbE_{\vc,\xw_0, \xl_0} 
    \left[
    \log\sigma\left(r_\phi(\vc,\xw_0)-r_\phi(\vc,\xl_0)\right)\right]
    \label{eq:BT_reward_classiciation_loss}
\end{equation}
And in RLHF objective~\cref{equ:RLHF}, the uique global optimal solution $p^*_\theta$ takes the form:
\begin{equation}
    p^*_{\theta}(\x_0|\vc)=\pref(\x_0|\vc)\exp\left(r(\vc,\x_0)/\beta\right) / Z(\vc)
    \label{eq:dpo_p_optimal}
\end{equation}
where $Z(\vc)=\sum_{\x_0}\pref(\x_0|\vc)\exp\left(r(\vc,\x_0)/\beta\right)$ is the partition function.
Hence, the reward fucntion is rewritten as 
\begin{equation}\label{eq:orig-dpo-r}
    r(\vc,\x_0) = \beta \log \frac{p^*_{\theta}(\x_0|\vc)}{\pref(\x_0|\vc)} + \beta\log Z(\vc)
\end{equation}
Using~\cref{eq:BT_reward_classiciation_loss} the reward objective becomes:
\begin{equation}
    L_{\text{DPO}}(\theta)\!=\!-
    \bbE_{\vc,\xw_{0},\xl_0}\left[
    \log\sigma\left(\beta \log \frac{p_{\theta}(\xw_0|\vc)}{\pref(\xw_0|\vc)}-\beta \log \frac{p_{\theta}(\xl_0|\vc)}{\pref(\xl_0|\vc)}\right)\right]
    \label{eq:dpo_loss_lang}
\end{equation}
By this reparameterization, instead of optimizing the reward function $r_\phi$ and then performing RL, \cite{rafailov2023direct} directly optimizes the optimal conditional distribution $p_\theta(\x_0|\vc)$.

\noindent\textbf{Diffusion DPO} \ \ Diffusion DPO extend the DPO formulation to diffusion models by defining the reward function over full denoising trajectories. Specifically, they assume that the reward is given by the expected trajectory-level score:

\begin{equation}
r(\vc,\x_0)=\bbE_{p_\theta(\x_{1:T}|\x_0,\vc)}\left[R(\vc,\x_{0:T})\right].
    \label{eq:define-r-all-t}
\end{equation}
Under this definition, the RLHF objective from~\cref{equ:RLHF} can be reformulated as:
\begin{align}
    \max_{p_\theta} \bbE_{\vc \sim \calD_c, \x_{0:T}\sim p_\theta(\x_{0:T}|\vc)} \left[r(\vc,\x_0)\right] - \beta \kl\left[p_\theta(\x_{0:T}|\vc)\|p_{\text{ref}}(\x_{0:T}|\vc)\right].
    \label{eq:dpo-objective-R}
\end{align}
Similar with~\cref{eq:dpo_p_optimal}, the optimal policy $p_\theta^*(\x_{0:T}|\vc)$ has a unique closed-form solution:
\begin{align}
    p_\theta^*(\x_{0:T}|\vc) = \pref(\x_{0:T}|\vc) \exp(R(\vc,\x_{0:T})/\beta) / Z(\vc),
\end{align}
where $Z(\vc)=\sum_{\x}\pref(\x_{0:T}|\vc)\exp\left(r(\vc,\x_0)/\beta\right)$ is the partition function.
The reparameterized form of the trajectory-level reward is given by
\begin{align}
    R(\vc, \x_{0:T}) & = \beta \log\frac{p_\theta^*(\x_{0:T}|\vc)}{\pref(\x_{0:T}|\vc)} + \beta \log Z(\vc).
\end{align}
Inserting this expression into the definition of $r(\vc, \x_0)$~\cref{eq:define-r-all-t} leads to 
\begin{equation*}
r(\vc,\x_0)=\beta \bbE_{p_\theta(\x_{1:T}|\x_0,\vc)}\left[ \log \frac{p^*_{\theta}(\x_{0:T}|\vc)}{\pref(\x_{0:T}|\vc)}\right] + \beta\log Z(\vc).
\end{equation*}
Substituting the resulting reward into the Bradley–Terry objective~\cref{eq:BT_reward_classiciation_loss} yields a maximum likelihood formulation for diffusion models. Since the partition function cancels in the pairwise setting, the per-example loss becomes:
\begin{equation}
\begin{aligned}
L_\text{D-DPO}(\theta)
&=
-\log \sigma \Biggl(
\beta \,
{\mathbb{E}}_{{\xw_{1:T} \sim p_\theta(\x_{1:T}\mid\xw_0,\vc), \,
\xl_{1:T} \sim p_\theta(\xl_{1:T}\mid\xl_0,\vc)}} \\
&\qquad \left[
\log \frac{p_{\theta}(\xw_{0:T} \mid \vc)}{\pref(\xw_{0:T} \mid \vc)}
-
\log \frac{p_{\theta}(\xl_{0:T} \mid \vc)}{\pref(\xl_{0:T}\mid \vc)}
\right]
\Biggr)
\end{aligned}
\end{equation}
Here, $\x^w_0$ and $\x^l_0$ are drwan from a static dataset, and the conditioning input $\vc$ is omitted for clarity.

To address the intractability of directly sampling from the true reverse process \( p_\theta(\x_{1:T} \mid \x_0) \), the Diffusion DPO formulation approximates it using the forward posterior \( q(\x_{1:T} \mid \x_0) \). Under this approximation, the trajectory-level preference objective can be rewritten as an expectation over individual timesteps. Specifically, comparisons between full trajectories are expressed as averages over comparisons between corresponding reverse transitions at each timestep, yielding the following expression:
\begin{align}
\begin{split}\label{eq-app:l1}
L_\text{D-DPO}(\theta) 
& = -\log \sigma \Biggl( \beta T 
     \bbE_t \, {\mathbb{E}}_{{\xw_{t-1,t} \sim q(\x_{t-1,t}|\xw_0),\;
     \xl_{t-1,t} \sim q(\x_{t-1,t}|\xl_0)}}  \\
& \quad \left[ \log \frac{p_{\theta}(\xw_{t-1}|\xw_t,\vc)}{\pref(\xw_{t-1}|\xw_t,\vc)} 
 - \log \frac{p_{\theta}(\xl_{t-1}|\xl_t,\vc)}{\pref(\xl_{t-1}|\xl_t,\vc)} 
\right] \Biggr)
\end{split}
\end{align}
To derive a tractable approximation, Jensen’s inequality is applied to move the logarithm outside the expectation, yielding an upper bound on the original loss:
\begin{align}
\begin{split}
\label{eq:D-DPO_kld}
L_\text{D-DPO}(\theta)
& \leq 
- \mathbb{E}_{\substack{
t,\, \xw_t \sim q(\x_t|\xw_0),\\
\xl_t \sim q(\x_t|\xl_0)
}} 
    \log \sigma \Biggl( \beta T \cdot \\
& \quad \mathbb{E}_{\substack{
\xw_{t-1} \sim q(\x_{t-1}|\xw_t,\xw_0),\\
\xl_{t-1} \sim q(\x_{t-1}|\xl_t,\xl_0)
}} 
    \Bigl[ 
        \log \frac{p_{\theta}(\xw_{t-1}|\xw_t,\vc)}{\pref(\xw_{t-1}|\xw_t,\vc)} 
        - \log \frac{p_{\theta}(\xl_{t-1}|\xl_t,\vc)}{\pref(\xl_{t-1}|\xl_t,\vc)} 
    \Bigr] 
    \Biggr) \\
& =
- \mathbb{E}_{\substack{
t,\, \xw_t \sim q(\x_t|\xw_0),\\
\xl_t \sim q(\x_t|\xl_0)
}} 
    \log \sigma \Biggl( \beta T \cdot \\
& \quad \mathbb{E}_{\substack{
\xw_{t-1} \sim q(\x_{t-1}|\xw_t,\xw_0),\\
\xl_{t-1} \sim q(\x_{t-1}|\xl_t,\xl_0)
}} 
    \Bigl[ 
        \log \frac{p_{\theta}(\xw_{t-1}|\xw_t,\vc)}{q(\xw_{t-1}|\xw_{0,t})} 
        - \log \frac{q(\xw_{t-1}|\xw_{0,t})}{\pref(\xw_{t-1}|\xw_t,\vc)} \\
& \qquad -\Bigl( \log \frac{p_{\theta}(\xl_{t-1}|\xl_t,\vc)}{q(\xl_{t-1}|\xl_{0,t})} 
        - \log \frac{q(\xl_{t-1}|\xl_{0,t})}{\pref(\xl_{t-1}|\xl_t,\vc)} \Bigr)
    \Bigr] 
    \Biggr) \\
& = 
- \mathbb{E}_{\substack{
t,\, \xw_t \sim q(\x_t|\xw_0),\\
\xl_t \sim q(\x_t|\xl_0)
}} 
    \log \sigma \Biggl( -\beta T \cdot \Bigl( \\
& \qquad \kl\left(q(\xw_{t-1}|\xw_{0,t}) \,\|\, p_\theta(\xw_{t-1}|\xw_t,\vc)\right) \\
& \quad
- \kl\left(q(\xw_{t-1}|\xw_{0,t}) \,\|\, \pref(\xw_{t-1}|\xw_t,\vc)\right) \\
& \quad - \Bigl( \kl\left(q(\xl_{t-1}|\xl_{0,t}) \,\|\, p_\theta(\xl_{t-1}|\xl_t,\vc)\right) \\
& \quad
- \kl\left(q(\xl_{t-1}|\xl_{0,t}) \,\|\, \pref(\xl_{t-1}|\xl_t,\vc)\right) \Bigr)
\Bigr) \Biggr)
\end{split}
\end{align}
Using the Gaussian parameterization of the reverse process (\cref{eq:reverse_likelihood}), the above loss simplifies to:
\begin{align}
\begin{split}
    L_{\text{D-DPO}}(\theta)
    &\leq - \mathbb{E}_{(\x_0^w, \x_0^l) \sim \mathcal{D}, t \sim \mathcal{U}(0, T), \x_t^w \sim q(\x_t^w|\x_0^w), \x_t^l \sim q(\x_t^l|\x_0^l) } 
\log \sigma \bigg( - \beta \cdot \big[ \nonumber \\
& \quad \| \noise^w - \noise_\theta(\x_t^w, t, \vc) \|_2^2 
- \| \noise^w - \noise_{\text{ref}}(\x_t^w, t, \vc) \|_2^2 \\
& \quad
- (\| \noise^l - \noise_\theta(\x_t^l, t, \vc) \|_2^2 
- \| \noise^l - \noise_{\text{ref}}(\x_t^l, t, \vc) \|_2^2)
\big] \bigg)
\end{split}
\end{align}
where $\noise^{w}, \noise^{l} \sim \calN(0,I)$, $\x_{t} \sim q(\x_t|\x_0)$ thus $\x_t = \alpha_t\x_0+\sigma_t\noise$. $\lambda_t=\alpha_t^2/\sigma_t^2$ is a signal-to-noise ratio term~\cite{vaediffusion}, in practice, the reweighting assigns each term the same weight~\cite{ddpm}.

\noindent\textbf{DDSPO} \ \
The DDSPO objective reformulates preference supervision in the diffusion framework by focusing on intermediate denoising transitions rather than final samples. Unlike standard DPO, which defines preferences over terminal outputs \( (\x_0^w, \x_0^l \mid \vc) \), the DDSPO formulation assumes supervision over denoising transitions \( ((\x_t^w, \x_{t-1}^w), (\x_t^l, \x_{t-1}^l) \mid \vc) \) at arbitrary timesteps \( t \).

Let \( p_\star^w(\x_{t-1,t} \mid \vc) \) and \( p_\star^l(\x_{t-1,t} \mid \vc) \) denote preferred and dispreferred denoising distributions respectively, conditioned on prompt \(\vc\). Following the Bradley–Terry approach in \cref{eq:BT_reward_classiciation_loss}, the reward is redefined over denoising transitions as a function \( r(\vc, \x_t, \x_{t-1}) \), with preference assigned to transitions sampled from \( p_\star^w \) over those from \( p_\star^l \). 

In this formulation, preference over transitions can be modeled analogously to \cref{equ:BT_distribution}, by replacing the reward function over final samples \( r(c, x_0) \) with a reward function over transitions \( r(c, x_t, x_{t-1}) \):
\begin{equation}
\mathbb{P}((\x_t^w, \x_{t-1}^w) \succ (\x_t^l, \x_{t-1}^l) \mid \vc) = \sigma\left(r(\vc, \x_t^w, \x_{t-1}^w) - r(\vc, \x_t^l, \x_{t-1}^l)\right),
\label{equ:BT_distribution_transition}
\end{equation}

Similar to \cref{eq:BT_reward_classiciation_loss}, the reward model \( r(\vc, \x_t, \x_{t-1}) \) can be parameterized by a neural network \( \phi \), and trained using maximum likelihood estimation for binary classification:
\begin{equation}
    L_\text{BT}(\phi) = - 
    \mathbb{E}_{\vc, \xw_{t-1}, \xw_{t}, \xl_{t-1}, \xl_{t}} 
    \left[
        \log \sigma\left(r_\phi(\vc, \xw_{t}, \xw_{t-1}) - r_\phi(\vc, \xl_{t}, \xl_{t-1})\right)
    \right]
    \label{eq:BT_reward_classiciation_loss_ddspo}
\end{equation}
This contrasts with standard DPO or Diffusion DPO, where preference pairs \((\x_0^w, \x_0^l \mid \vc)\) are drawn from a dataset \(\mathcal{D}\) of full samples. In DDSPO, supervision instead arises from tuples \(((\x_t^w, \x_{t-1}^w), (\x_t^l, \x_{t-1}^l) \mid \vc)\) sampled from preferred and dispreferred denoising policies $p_\star^w(\x_{t-1,t} \mid \vc)$ and $p_\star^l(\x_{t-1,t} \mid \vc)$. This allows DDSPO to directly optimize the model's denoising behavior at each timestep \(t\), rather than the distribution over final outputs.
This formulation enables a direct comparison between denoising transitions at the same timestep, providing fine-grained supervision over the reverse process.

Accordingly, the DDSPO objective adopts a DPO-style structure defined over transitions, analogous to~\cref{equ:RLHF}:
\begin{multline}
    \max_{p_\theta} \bbE_{\vc,\, \x_{t-1,t} \sim p_\theta(\x_{t-1,t} \mid \vc)} 
    \left[ r(\vc, \x_t, \x_{t-1}) \right] \\
    - \beta \, \kl \left[ p_\theta(\x_{t-1}|\x_t, \vc) \, \| \, p_{\text{ref}}(\x_{t-1}|\x_t, \vc) \right]
    \label{eq:ddspo_objective}
\end{multline}
The optimal reverse transition distribution under this formulation has the form:
\begin{equation}
    p^*_{\theta}(\x_{t-1}|\x_t, \vc)
    = \pref(\x_{t-1}|\x_t, \vc) \cdot 
    \frac{\exp(r(\vc, \x_t, \x_{t-1}) / \beta)}{Z(\vc)},
    \label{eq:ddspo_optimal_policy}
\end{equation}
where the partition function  $Z(\vc) $ is defined as
\[
Z(\vc) = \sum_{\x_t, \x_{t-1}} \pref(\x_{t-1}|\x_t, \vc) \exp(r(\vc, \x_t, \x_{t-1}) / \beta).
\]
The reward can thus be reparameterized as
\begin{equation}
    r(\vc, \x_t, \x_{t-1}) 
    = \beta \log \frac{p^*_\theta(\x_{t-1}|\x_t, \vc)}{\pref(\x_{t-1}|\x_t, \vc)} + \beta \log Z(\vc).
    \label{eq:ddspo_reward_reparam}
\end{equation}
Substituting the reward expression from~\cref{eq:ddspo_reward_reparam} into the Bradley--Terry objective~\cref{eq:BT_reward_classiciation_loss_ddspo} yields a per-example loss defined over denoising transitions:
\begin{equation}
\begin{aligned}
\ell_{\text{DDSPO}}^{(t)}(\theta) = 
\mathbb{E}_{\substack{
\vc \sim \mathcal{D}(c),\,
(\x_{t-1}^w, \x_t^w) \sim p_\star^w(\x^w_{t-1,t} \mid \vc),\\
(\x_{t-1}^l, \x_t^l) \sim p_\star^l(\x^l_{t-1,t} \mid \vc)
}}
\Bigl[
- \log \sigma \Bigl( \\
\beta \log \frac{
p_{\theta}(\x_{t-1}^w \mid \x_t^w, c)
}{
p_{\text{ref}}(\x_{t-1}^w \mid \x_t^w, c)
}
-
\beta \log \frac{
p_{\theta}(\x_{t-1}^l \mid \x_t^l, c)
}{
p_{\text{ref}}(\x_{t-1}^l \mid \x_t^l, c)
}
\Bigr)
\Bigl]
\end{aligned}
\end{equation}
The full DDSPO objective is then obtained by taking the expectation over uniformly sampled timesteps \( t \sim \mathcal{U}(0, T) \). 
Equivalently, this corresponds to uniformly averaging per-timestep losses across all diffusion steps:
\begin{equation}
\mathcal{L}_{\text{DDSPO}}(\theta) = 
\mathbb{E}_{t \sim \mathcal{U}(0, T)} \left[ \ell_{\text{DDSPO}}^{(t)}(\theta) \right],
\end{equation}
Concretely, we instantiate it as:
\begin{multline*}
\mathcal{L}_{\text{DDSPO}}(\theta) = 
 - \mathbb{E}_{\vc\sim \mathcal{D}(c),\,t \sim \mathcal{U}(0, T), \,
(\x_{t-1}^w, \x_t^w) \sim p_\star^w(\x^w_{t-1,t} \mid \vc),
(\x_{t-1}^l, \x_t^l) \sim p_\star^l(\x^l_{t-1,t} \mid \vc)
} \\
\log\sigma\!\left(
\beta \log\!\frac{p_{\theta}(\x_{t-1}^w\mid\x_t^w,\vc)}{p_{\text{ref}}(\x_{t-1}^w\mid\x_t^w,\vc)}
-\beta \log\!\frac{p_{\theta}(\x_{t-1}^l\mid\x_t^l,\vc)}{p_{\text{ref}}(\x_{t-1}^l\mid\x_t^l,\vc)}
\right)
\end{multline*}
As in Diffusion DPO, we approximate the intractable joint transition distribution
\begin{align}
\label{eq:DDSPO_approx}
p_\star^w(\x_{t-1,t}^w \mid c) \approx q(\x_t^w \mid \x_0^w) \cdot p_\star(\x_{t-1}^w \mid \x_t^w, c),
\end{align}
where \( \x_0^w \sim \mathcal{D} \), \( \x_t^w \sim q(\x_t \mid \x_0^w) \), and \( \x_{t-1}^w \sim p_\star(\x_{t-1} \mid \x_t^w, c) \).
The same applies to the dispreferred trajectory \( (\x_t^l, \x_{t-1}^l) \).

Substituting this into the DDSPO objective yields:
\begin{align}
\begin{split}
\mathcal{L}_{\text{DDSPO}}(\theta) 
&=
\mathbb{E}_{\substack{
\vc \sim \mathcal{D}(c),\,
t \sim \mathcal{U}(0, T),\\
(\x_{t-1}^w, \x_t^w) \sim q(\x_t^w \mid \x_0^w)\cdot p_\star^w(\x_{t-1}^w \mid \x_t^w, c),\\
(\x_{t-1}^l, \x_t^l) \sim q(\x_t^l \mid \x_0^l)\cdot p_\star^l(\x_{t-1}^l \mid \x_t^l, c)
}}
\Bigl[
- \log \sigma \Bigl(
\\
&\qquad
\beta \log \frac{
p_{\theta}(\x_{t-1}^w \mid \x_t^w, c)
}{
p_{\text{ref}}(\x_{t-1}^w \mid \x_t^w, c)
}
-
\beta \log \frac{
p_{\theta}(\x_{t-1}^l \mid \x_t^l, c)
}{
p_{\text{ref}}(\x_{t-1}^l \mid \x_t^l, c)
}
\Bigr)
\Bigr]
\\
& \leq 
- \mathbb{E}_{\substack{
\vc \sim \mathcal{D}(c),\,
t \sim \mathcal{U}(0, T),\\
\xw_t \sim q(\x_t|\xw_0),\,
\xl_t \sim q(\x_t|\xl_0)
}}
\log \sigma \Biggl( \beta T \cdot \\
& \quad \mathbb{E}_{\substack{
\xw_{t-1} \sim p_\star^w(\xw_{t-1}|\xw_t, \vc),\\
\xl_{t-1} \sim p_\star^l(\xl_{t-1}|\xl_t, \vc)
}}
\Bigl[
\log \frac{p_{\theta}(\xw_{t-1}|\xw_t,\vc)}{\pref(\xw_{t-1}|\xw_t,\vc)} 
- \log \frac{p_{\theta}(\xl_{t-1}|\xl_t,\vc)}{\pref(\xl_{t-1}|\xl_t,\vc)} 
\Bigr]
\Biggr) \\
& =
- \mathbb{E}_{\substack{
\vc \sim \mathcal{D}(c),\,
t \sim \mathcal{U}(0, T),\\
\xw_t \sim q(\x_t|\xw_0),\,
\xl_t \sim q(\x_t|\xl_0)
}}
\log \sigma \Biggl( \beta T \cdot \\
& \quad \mathbb{E}_{\substack{
\xw_{t-1} \sim p_\star^w(\xw_{t-1}|\xw_t, \vc),\\
\xl_{t-1} \sim p_\star^l(\xl_{t-1}|\xl_t, \vc)
}}
\Bigl[
\log \frac{p_{\theta}(\xw_{t-1}|\xw_t,\vc)}{p_\star^w(\xw_{t-1}|\xw_t,\vc)} 
- \log \frac{p_\star^w(\xw_{t-1}|\xw_t,\vc)}{\pref(\xw_{t-1}|\xw_t,\vc)}
\\
& \qquad
-\Bigl(
\log \frac{p_{\theta}(\xl_{t-1}|\xl_t,\vc)}{p_\star^l(\xl_{t-1}|\xl_t,\vc)} 
- \log \frac{p_\star^l(\xl_{t-1}|\xl_t,\vc)}{\pref(\xl_{t-1}|\xl_t,\vc)}
\Bigr)
\Bigr]
\Biggr) \\
& = 
- \mathbb{E}_{\substack{
\vc \sim \mathcal{D}(c),\,
t \sim \mathcal{U}(0, T),\\
\xw_t \sim q(\x_t|\xw_0),\,
\xl_t \sim q(\x_t|\xl_0)
}}
\log \sigma \Biggl( -\beta T \cdot \Bigl( \\
& \qquad
\kl\left(p_\star^w(\xw_{t-1}|\xw_t,\vc) \,\|\, p_\theta(\xw_{t-1}|\xw_t,\vc)\right) \\
& \quad
- \kl\left(p_\star^w(\xw_{t-1}|\xw_t,\vc) \,\|\, \pref(\xw_{t-1}|\xw_t,\vc)\right) \\
& \quad
- \Bigl(
\kl\left(p_\star^l(\xl_{t-1}|\xl_t,\vc) \,\|\, p_\theta(\xl_{t-1}|\xl_t, \vc)\right) \\
& \quad
- \kl\left(p_\star^l(\xl_{t-1}|\xl_t,\vc) \,\|\, \pref(\xl_{t-1}|\xl_t,\vc)\right)
\Bigr)
\Bigr) \Biggr)
\end{split}
\end{align}
Using the Gaussian parameterization of the reverse process~\cref{eq:reverse_likelihood}, the above loss simplifies to:
\begin{align}
\label{eq:DDSPO_sup}
\mathcal{L}_{\text{DDSPO}}&(\theta) \leq 
 - \mathbb{E}_{\substack{
(\x_0^w,\x_0^l) \sim \mathcal{D},\,\vc\sim \mathcal{D}(c), t \sim \mathcal{U}(0, T), \\ \x_t^w \sim q(\x_t \mid \x_0^w), \x_t^l \sim q(\x_t \mid \x_0^l)
}} 
\log \sigma \bigg( - \beta \cdot \big[ \notag \\
& \qquad \| \noise_{\star}^w - \noise_\theta(\x_t^w, t, \vc) \|_2^2 
- \| \noise_{\star}^w - \noise_{\text{ref}}(\x_t^w, t, \vc) \|_2^2 \notag \\
& \quad - \big( 
\| \noise_{\star}^l - \noise_\theta(\x_t^l, t, \vc) \|_2^2 
- \| \noise_{\star}^l - \noise_{\text{ref}}(\x_t^l, t, \vc) \|_2^2 
\big) 
\big] \bigg).
\end{align}

\noindent\textbf{DSPO} \ \
DSPO~\cite{zhu2025dspo} follows the standard score decomposition used in conditional diffusion models and expresses the conditional score as
\begin{equation}
\label{eq:score_decomp_appendix}
\nabla_{\x_t^w} \log p_\theta(\x_t^w \mid \vc, y)
=
\nabla_{\x_t^w} \log p_\theta(\x_t^w \mid \vc)
+
\nabla_{\x_t^w} \log p(y \mid \x_t^w, \vc),
\end{equation}
where $y$ represents a human-preference condition.
DSPO proposes to fine-tune the pretrained score model
to match a preference-adjusted target score, leading to the objective
\begin{equation}
\label{eq:appendix_dspo_objective}
\min_\theta \;
\omega(t)
\left\|
\nabla_{\x_t^w} \log p_\theta(\x_t^w \mid \vc)
-
\left(
\nabla_{\x_t^w} \log p(\x_t^w \mid \vc)
+
\gamma \nabla_{\x_t^w} \log p(y \mid \x_t^w, \vc)
\right)
\right\|_2^2 ,
\end{equation}
where $\omega(t)$ is the diffusion timetable weighting function
and $\gamma$ scales the preference constraint.
The preference likelihood is modeled as
\begin{equation}
\label{eq:appendix_pref_prob}
p(y \mid \x_t^w, \vc)
=
p(\x_t^w \succ \x_t^l \mid \x_t^l, \vc)
=
\sigma\left(
r(\x_t^w, t, \vc) - r(\x_t^l, t, \vc)
\right).
\end{equation}
By substituting this likelihood into \cref{eq:appendix_dspo_objective} and
deriving the resulting score-matching objective,
the loss can be extended over all timesteps to yield:
\begin{multline}
\mathcal{L}_{\text{DSPO}}(\theta)
= \mathbb{E}_{\substack{
(\x_0^w,\x_0^l)\sim \mathcal{D}, \, \vc\sim \mathcal{D}(c), \, t \sim \mathcal{U}(0,T), \\ \x_t^w \sim q(\x_t^w \mid \x_0^w), \, \x_t^l \sim q(\x_t^l \mid \x_0^l)
}}
A(t)
\Big\|
(\noise_\theta(\x_t^w,t,\vc) - \noise^w)
\\
\quad
- \beta
\Big(
1 - \sigma\big(
r_\theta(\x_t^w,t,\vc) - r_\theta(\x_t^l,t,\vc)
\big)
\Big)
\cdot
(\noise_\theta(\x_t^w,t,\vc) - \noise_\text{ref}(\x_t^w,t,\vc))
\Big\|_2^2.
\end{multline}
Here, $A(t)$ denotes a timestep-dependent weighting function; following DSPO, we disregard $A(t)$ and the associated parameters involving $\alpha_t$ and $\beta_t$ at the beginning of the objective.
In DSPO, to avoid training an additional probability model,
$r(\x_t, \vc)$ is computed using the implicit reward formulation of
Diffusion-DPO~\cite{Wallace_2024_CVPR}
\begin{equation}
\label{eq:appendix_reward_original}
r(\x_t, t, \vc)
=
\lambda
\log
\frac{p_\theta(\x_{t-1} \mid \x_{t}, \vc)}
     {p_{\text{ref}}(\x_{t-1} \mid \x_{t}, \vc)},
\end{equation}
DSPO approximates $\x_t$ using its posterior mean under the forward process as
\[
\mathbb{E}[\x_{t-1} \mid \x_{t}, x_0]
=
\sqrt{\frac{\alpha_{t-1}}{\alpha_{t}}}
\left(
\x_{t}
-
\frac{\beta_{t}}{\sqrt{1-\bar{\alpha}_{t}}}\,\boldsymbol{\epsilon}
\right).
\]
Then, using this approximation, the estimated $p_\theta(\x_{t-1} \mid \x_{t}, c)$ becomes
\[
\lambda
\left(\sqrt{2\pi\sigma_{t}^2}\right)^{-d}
\exp\left(
-\frac{1}{2}
\frac{\beta_{t}}{1-\bar{\alpha}_{t-1}}
\frac{\alpha_{t-1}}{\alpha_{t}}
\left\|
\noise
-
\noise_\theta(\x_{t}, t, \vc)
\right\|_2^2
\right).
\]
Therefore, the reward function
$r(\cdot) = \lambda (\log p_\theta(\x_{t-1} \mid \x_{t}, \vc) - \log p_{\text{ref}}(\x_{t-1} \mid \x_{t}, \vc))$
is represented in DSPO as
\[
r(\x_t, t, \vc)
=
-\frac{\lambda}{2}
\frac{\beta_{t}}{1-\bar{\alpha}_{t-1}}
\frac{\alpha_{t-1}}{\alpha_{t}}
\left(
\left\|
\noise
-
\noise_\theta(\x_{t}, t, \vc)
\right\|_2^2
-
\left\|
\noise
-
\noise_{\text{ref}}(\x_{t}, t, \vc)
\right\|_2^2
\right).
\]

In contrast, our formulation avoids this forward-process approximation and
derives the denoising target directly from the contrastive denoising policies.
Specifically, instead of using
$\boldsymbol{\epsilon}$ obtained from the forward posterior mean approximation, we incorporate
a policy-grounded target $\boldsymbol{\epsilon}_\star$ extracted from the winning and losing
denoising policies $p_\star^w(\x_{t-1}^w \mid \x_{t}^w, \vc)$ and
$p_\star^l(\x_{t-1}^l \mid \x_{t}^l, \vc)$, respectively. This leads to a modified approximation
of the form
\[
\mathbb{E}[\x_{t-1} \mid \x_{t}, x_0]
=
\sqrt{\frac{\alpha_{t-1}}{\alpha_{t}}}
\left(
\x_{t}
-
\frac{\beta_{t}}{\sqrt{1-\bar{\alpha}_{t}}}\,\boldsymbol{\epsilon}_\star
\right),
\]
and accordingly, the reward expression becomes
\[
r(\x_t, t, \vc)
=
-\frac{\lambda}{2}
\frac{\beta_{t}}{1-\bar{\alpha}_{t-1}}
\frac{\alpha_{t-1}}{\alpha_{t}}
\left(
\left\|
\noise_\star
-
\noise_\theta(\x_{t}, t, \vc)
\right\|_2^2
-
\left\|
\noise_\star
-
\noise_{\text{ref}}(\x_{t}, t, \vc)
\right\|_2^2
\right).
\]
Through this replacement, the reward is directly derived from the contrastive
denoising policies rather than from a forward-process approximation, enabling
policy-grounded preference supervision at each denoising step.

\section{Extended Experimental Results}
\label{sec:supp_extended}
In this section, we provide extended results and additional evaluations
omitted from the main paper due to space constraints.

\begin{figure}[t]
    \centering
    \includegraphics[width=1\linewidth]{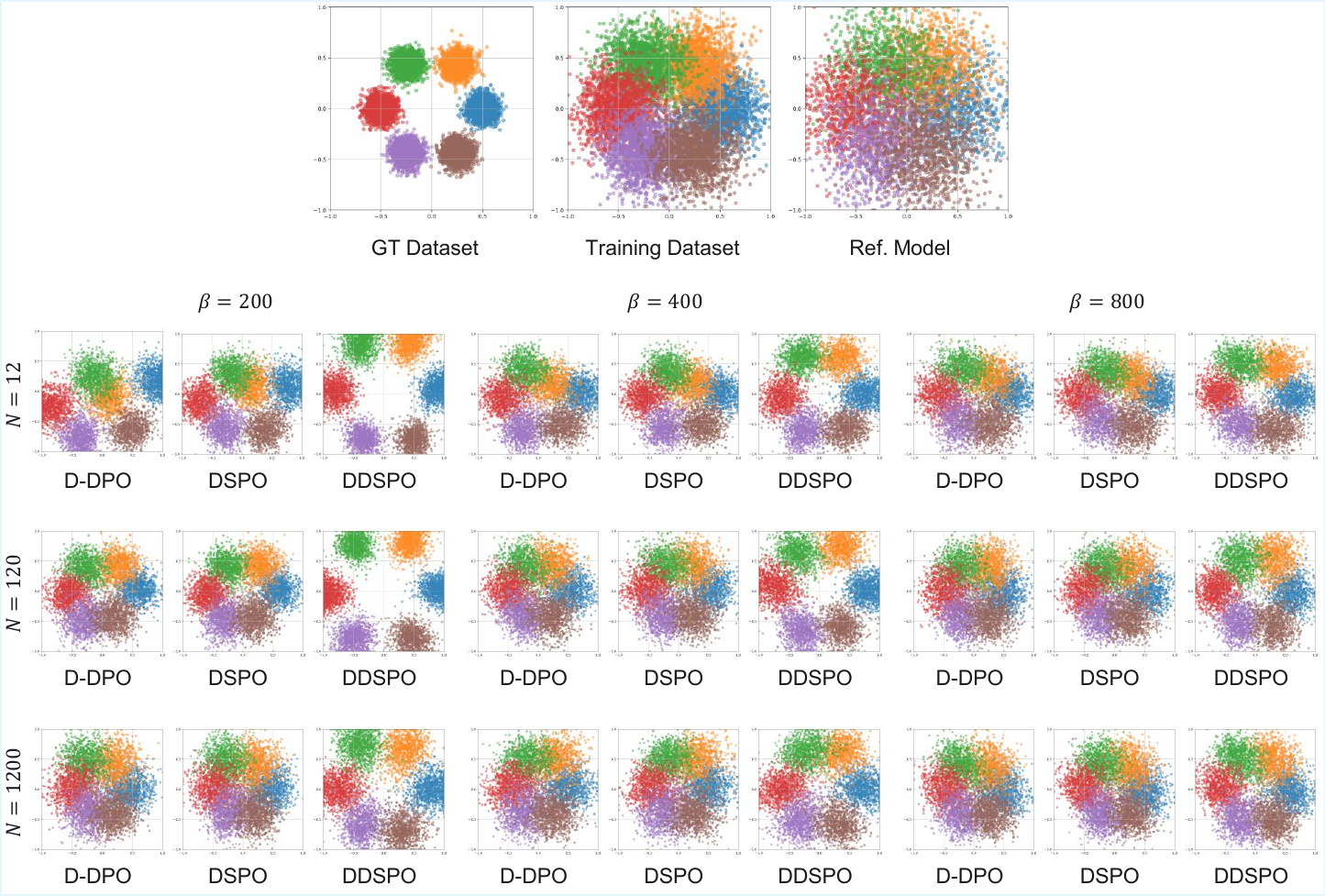}
    \caption{\textbf{Full Results of Toy Experiments for Comparison of Diffusion DPO (D-DPO), DSPO and DDSPO.} Qualitative results under varying dataset size ($N \in \{12, 120, 1200\}$) and regularization strengths ($\beta \in \{200, 400, 800\}$).}
    \label{fig:sub_toy_full}
\end{figure}

\noindent\textbf{Toy Experiment Results} \ \
We report extended results from the 2D toy experiment introduced in \Cref{sec:exp:toy}. Specifically, we compare Diffusion DPO and DDSPO across various settings by varying the dataset size and the hyperparameter $\beta$. \cref{fig:sub_toy_full} presents results across different combinations of preference pair counts ($N \in {12, 120, 1200}$) and regularization strengths ($\beta \in {200, 400, 800}$).

Across all configurations, we observe that DDSPO consistently produces more well-separated and class-consistent outputs compared to Diffusion DPO. Both increasing the number of preference pairs per class ($N$) and the regularization strength ($\beta$) lead to improved separation for both methods, but these improvements are substantially more pronounced and stable with DDSPO. Notably, DDSPO maintains clear cluster boundaries and discriminative outputs even under low-data regimes (e.g., $N = 12$), whereas Diffusion DPO often suffers from mode overlap and distorted clusters, particularly when supervision is limited or hyperparameters are not carefully tuned. These results indicate that DDSPO is robust to the amount of supervision, consistently enabling the model to learn well-separated, condition-specific distributions by dynamically providing diverse guidance at each timestep and pushing samples away from dispreferred directions.

\begin{table}[t]
\centering
\caption{
\textbf{Comparison of Various Architectures and Methods on Alignment Benchmarks.}
We evaluate SD-1.4, SDXL, SANA and SD3-Medium models on GenEval and T2I-CompBench, each measuring alignment across six distinct aspects.
}
\label{tab:alignment_benchmark_results}
\vspace{-1em}

\begin{subtable}[t]{\linewidth}
\centering
\caption{GenEval}
\scalebox{0.85}{
\begin{tabular}{l ccccccc}
\hline
\hline
      & \multicolumn{7}{c}{GenEval}\\
      \cmidrule(lr){2-8}
Model & Single & Two & Counting & Colors & Position & Color Attribution & Overall \\
\hline



\rowcolor{gray!25} SD-1.4 & 0.98 & 0.35 & 0.32 & 0.73 & 0.05 & 0.07 & 0.4245\\

~+DDSPO  &1.00 & 0.53 & 0.44 & 0.81 & 0.12 & 0.13 & 0.5045\\
\hline
\rowcolor{gray!25} SDXL & 0.99 & 0.65 & 0.37 & 0.82 & 0.14 & 0.17 & 0.5229 \\

~+Itercomp & 1.00 & 0.83 & 0.43 & 0.86 & 0.22 & 0.33 & 0.6108 \\

~+CaPO & 0.99 & 0.79 & 0.48 & 0.86 & 0.15 & 0.28 & 0.5900\\

~+DDSPO & 1.00 & 0.88 & 0.41 & 0.90 & 0.20 & 0.25 & 0.6049\\
\hline

\rowcolor{gray!25} SANA & 0.99 & 0.79 & 0.68 & 0.88 & 0.31 & 0.44 & 0.6812 \\
~+DDSPO & 1.00 & 0.88 & 0.69 & 0.90 & 0.37 & 0.52 & 0.7266 \\
\rowcolor{gray!25} SD3-M & 1.00 & 0.86 & 0.66 & 0.86 & 0.33 & 0.57 & 0.7123 \\
~+CaPO & 0.99 & 0.87 & 0.63 & 0.86 & 0.31 & 0.59 & 0.7100\\
~+DDSPO & 1.00 & 0.95 & 0.52 & 0.87 & 0.41 & 0.69 & 0.7384 \\

\hline
\hline
\end{tabular}
}

\end{subtable}

\vspace{+0.5em}

\begin{subtable}[t]{\linewidth}
\centering
\caption{T2I-Compbench}
\scalebox{0.85}{
\begin{tabular}{l cccccccc}
\hline
\hline
      & \multicolumn{8}{c}{T2I-Compbench}\\
      \cmidrule(lr){2-9}
Model & Color & Shape & Texture & Spatial & Non Spatial & Complex(B-VQA) & Overall & Complex(3-in-1)\\
\hline


\rowcolor{gray!25} SD-1.4 & 0.3593 & 0.3589 & 0.4024 & 0.1059 & 0.3096 & 0.3792 & 0.3192 & 0.3089\\

~+DDSPO &0.5123 & 0.4336 & 0.5425 & 0.1475 & 0.3130 &  0.4463 & 0.3992 & 0.3451 \\

\hline
\rowcolor{gray!25} SDXL & 0.5762 & 0.4774 & 0.5225 & 0.1969 & 0.3125 & 0.4253 & 0.4185& 0.3346\\

~+Itercomp & 0.7090 & 0.5254 & 0.6249 & 0.2400 & 0.3184 & 0.4899 & 0.4846 & 0.3687\\

~+CaPO & 0.6460 & 0.5370 & 0.6330 & 0.1720 & 0.3120 & 0.4910 & 0.4652 & - \\

~+DDSPO & 0.7658 & 0.5610 & 0.6631 & 0.2335 & 0.3175 & 0.4972 & 0.5064 & 0.3734\\
\hline

\rowcolor{gray!25} SANA & 0.6987 & 0.5330 & 0.6555 & 0.3212 & 0.3146 & 0.5176 & 0.5068 & 0.3844\\
~+DDSPO &0.7634 & 0.6001 & 0.7157 & 0.3545 & 0.3152 &0.5563  & 0.5509 & 0.4042 \\

\hline
\rowcolor{gray!25} SD3-M & 0.7986 & 0.5805 & 0.7192 & 0.3013 & 0.3141 & 0.5129 & 0.5378 & 0.3795 \\

~+CaPO & 0.7880 & 0.5720 & 0.7310 & 0.2300 & 0.3130 & 0.5090 & 0.5238 & - \\

~+DDSPO & 0.8641 & 0.6747 & 0.8129 & 0.3550 & 0.3206 & 0.5853 & 0.6021 & 0.4164\\

\hline
\hline
\end{tabular}
}

\end{subtable}
\end{table}

\noindent\textbf{Text-to-Image Alignment Results} \ \
As shown in Table~\ref{tab:alignment_benchmark_results}, our method consistently enhances the performance of all tested architectures across every subcategory of benchmarks, confirming its universal applicability and effectiveness.

In comparison to methods like IterComp and CaPO applied to the SDXL model, our method achieves superior results in the \textit{Two Objects} and \textit{Colors} categories of the GenEval benchmark, while slightly lagging behind in \textit{Counting} and \textit{Color Attribution}, which are also part of GenEval.
On the T2I-Compbench benchmark, our method outperforms existing approaches in \textit{Color}, \textit{Shape}, and \textit{Complex}, while showing slightly lower performance than IterComp in \textit{Spatial} and \textit{Non-Spatial}. Nonetheless, it achieves the highest \textit{Overall} score.

It is worth highlighting that, unlike IterComp and CaPO—both of which rely on reward models and human-annotated preference data—our method requires neither. Despite this, it achieves competitive results across both GenEval and T2I-CompBench, demonstrating that strong compositional alignment can be achieved with minimal supervision through self-consistency alone.

\begin{table}[t]
\centering
\caption{\textbf{Joint Improvement on Text-to-Image Alignment and Aesthetic Quality.} 
We evaluate DDSPO trained with different types of supervision. 
\textit{Rand-removal} uses 200K prompts with alignment-degraded negatives generated via random token removal. 
\textit{LLaMA (AQ)} uses 200K prompts with aesthetic-degraded negatives generated via LLaMA-3.
\textit{Rand-removal + LLaMA (AQ)} combines both sources by sampling 100K examples from each dataset (alignment and aesthetic), for a total of 200K.
All variants improve over the SDXL base model across both alignment and aesthetic metrics, demonstrating the flexibility of DDSPO across objectives.}
\scalebox{0.85}{
\begin{tabular}{lcccc}
    \hline
    \hline
    Method & \multicolumn{2}{c}{T2I Alignment} &  \multicolumn{2}{c}{Aesthetic Quality} \\
    \cmidrule(lr){2-3} \cmidrule(lr){4-5}&GenEval$\uparrow$ & CompBench$\uparrow$ & HPSv2$\uparrow$ & PickScore$\uparrow$ \\
    \hline
    \rowcolor{gray!25}
    SDXL & 0.5229 & 0.4034 & 27.89 & 22.27\\
    Rand-removal & 0.6049 & 0.4857 & 28.64 & 22.66 \\
    LLaMA (AQ) & 0.5808 & 0.4698 & 28.78 & 22.70 \\
    Rand-removal + LLaMA (AQ) & 0.6010& 0.4863 & 28.79 & 22.69 \\
    \hline
    \hline
\end{tabular}
}
\label{tab:sup_align_aesthetic}
\end{table}
\noindent\textbf{Improving Both Alignment and Aesthetic Quality}\ \
To evaluate the versatility of DDSPO, we conduct a joint experiment targeting both text-to-image alignment and aesthetic quality. Specifically, we reuse the datasets introduced in \cref{sec:exp:align,sec:exp:aesthetic}, each containing 200K training examples constructed for alignment and aesthetic supervision, respectively. For this joint setup, we randomly sample 100K examples from each and train DDSPO on the combined 200K preference pairs. All experiments are conducted by finetuning the base SDXL~\cite{SDXL} model.

We compare the jointly trained DDSPO model against the original SDXL model, as well as models trained using only alignment supervision (via random token removal) and only aesthetic supervision (via LLaMA-based degradation). As shown in \cref{tab:sup_align_aesthetic}, our jointly trained DDSPO model achieves consistent improvements in both \textit{T2I Alignment} (GenEval and CompBench) and \textit{Aesthetic Quality} (HPSv2 and PickScore), demonstrating the effectiveness of DDSPO across diverse objectives and its ability to integrate multiple supervision signals in a unified framework.

\begin{figure}[t]
    \centering
     \begin{tikzpicture}[xshift=+2em]
        \draw[line width=1.5pt, color=bkcolor] (-3em,0) -- ++(1em,0);
        \node[diamond, fill=bkcolor, scale=0.5] at (-1.8em,0) {};
        \node[font=\small, right=0.5em] at (-1em,0) {Diffusion DPO};

        \draw[line width=1.5pt, color=ourscolor] (8em,0) -- ++(1em,0);
        \node[circle, fill=ourscolor, scale=0.5] at (9.2em,0) {};
        \node[font=\small, right=0.5em] at (10em,0) {DDSPO};
    \end{tikzpicture}
    \vspace{0.5em}

    \begin{subfigure}[b]{0.35\textwidth}
        \includegraphics[width=\textwidth]{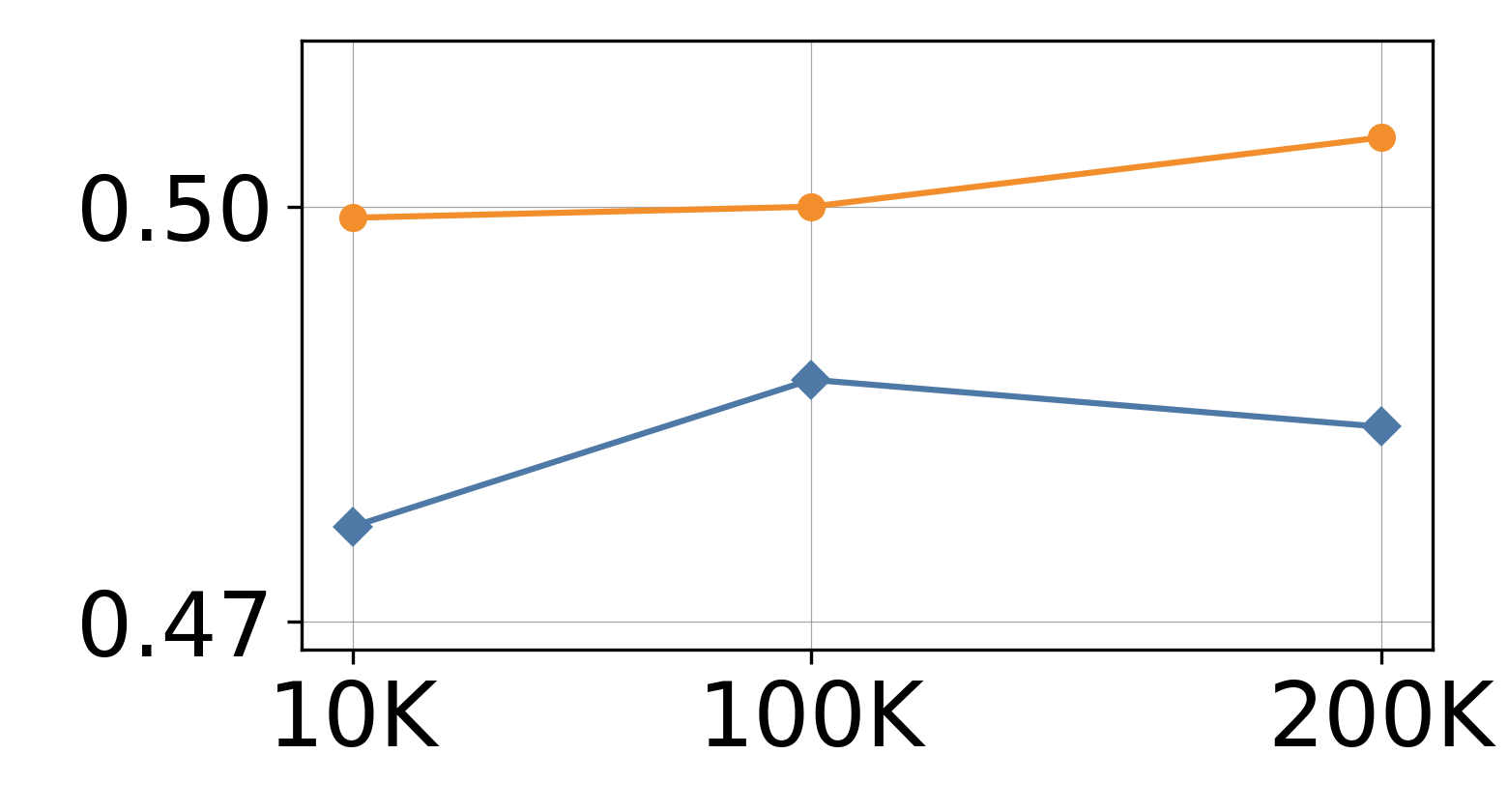}
        \caption{GenEval}
        \label{fig:sup:dataset_geneval}
    \end{subfigure}
    \begin{subfigure}[b]{0.35\textwidth}
        \includegraphics[width=\textwidth]{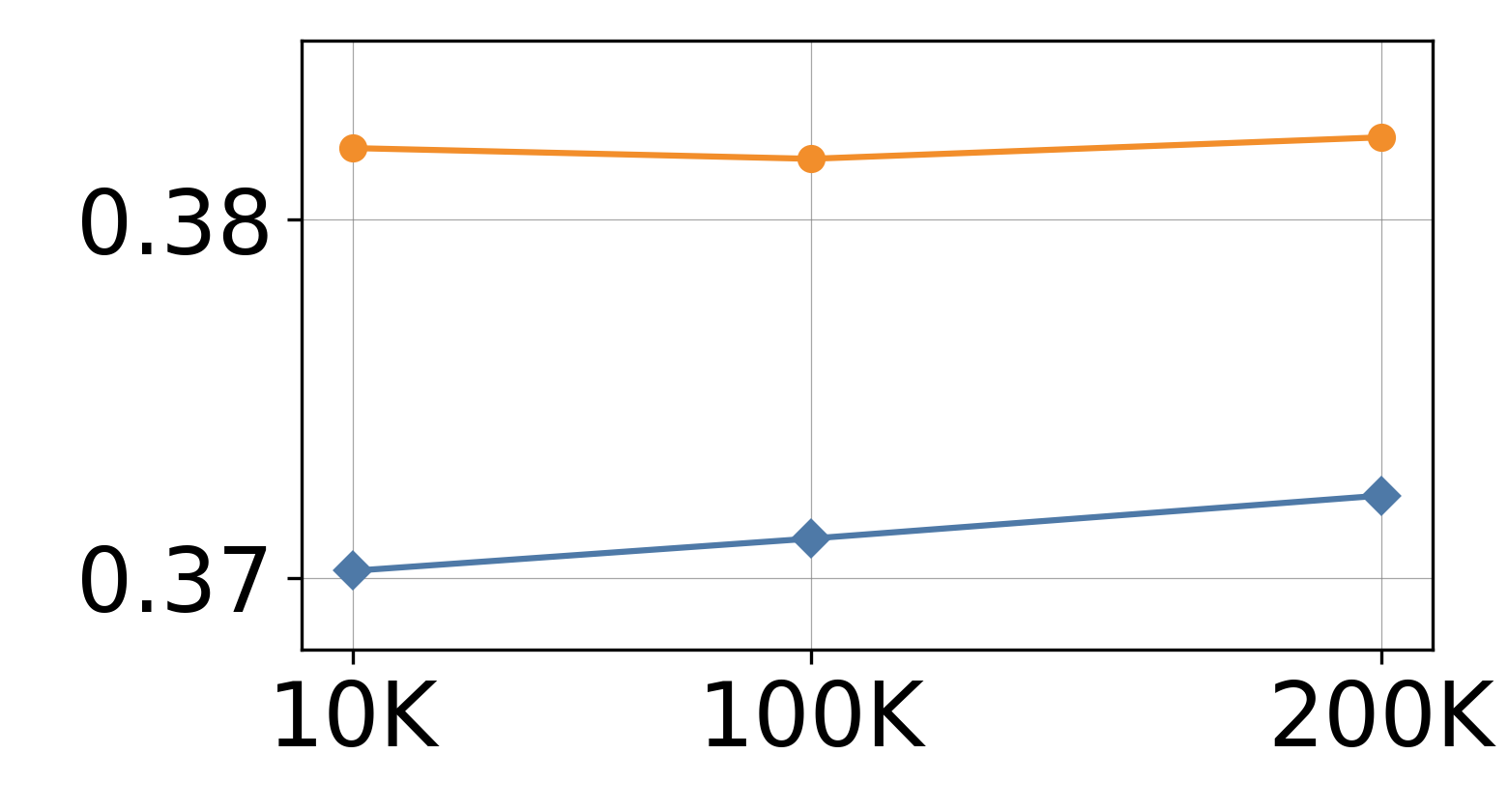}
        \caption{Compbench}
        \label{fig:sup:dataset_compbench}
    \end{subfigure}
    \caption{\textbf{Ablation Study on Dataset Size} Performance on GenEval (left) and Compbench (right) with various training set sizes. The orange line denotes our method, while the blue line denotes D-DPO. 
    }
    \label{fig:sup:dataset}
    
\end{figure}

\noindent\textbf{Dataset Size Ablation}\ \
We performed an ablation study to evaluate the effect of training dataset size on performance, using Stable Diffusion 1.4 as the base model. As illustrated in \cref{fig:sup:dataset}, we compare our method (orange) with D-DPO (blue) on the GenEval and CompBench benchmarks, using 10K, 100K, and 200K training pairs. The plotted values represent the overall alignment scores (averaged across metrics) for each benchmark.

Our method shows clear and consistent gains over D-DPO across all dataset sizes. Notably, even with only 10K preference pairs, our method achieves substantial improvements over the base model (whose GenEval and CompBench scores are 0.4245 and 0.3075, respectively), demonstrating its strong data efficiency. These results highlight the practicality and scalability of DDSPO, which remains highly effective even in low-data regimes.

\begin{table}[t]
\centering
\caption{
\textbf{Evaluation on SD3.5-Large.}
We compare the base SD3.5-Large model with DDSPO instantiated using
TF-CPP and report the full category-wise results on GenEval and
T2I-CompBench.
}
\label{tab:supp_sd35_large}
\vspace{-1em}

\begin{subtable}[t]{\linewidth}
\centering
\caption{GenEval}
\scalebox{0.85}{
\begin{tabular}{l ccccccc}
\hline
\hline
      & \multicolumn{7}{c}{GenEval}\\
      \cmidrule(lr){2-8}
Model
& Single Obj.
& Two Obj.
& Counting
& Colors
& Position
& Color Attr.
& Overall \\
\hline

\rowcolor{gray!25}
SD3.5-L
& 0.9910
& 0.9070
& 0.7160
& 0.8620
& 0.2630
& 0.5650
& 0.7170 \\

~+DDSPO
& 0.9970
& 0.9320
& 0.7280
& 0.8700
& 0.3050
& 0.6330
& 0.7440 \\

\hline
\hline
\end{tabular}
}
\end{subtable}

\vspace{0.5em}

\begin{subtable}[t]{\linewidth}
\centering
\caption{T2I-CompBench}
\scalebox{0.85}{
\begin{tabular}{l ccccccc}
\hline
\hline
      & \multicolumn{7}{c}{T2I-CompBench} \\
      \cmidrule(lr){2-8}
Model
& Color
& Shape
& Texture
& Spatial
& Non Spatial
& Complex
& Overall \\
\hline

\rowcolor{gray!25}
SD3.5-L
& 0.7760
& 0.5940
& 0.7230
& 0.2780
& 0.3170
& 0.3820
& 0.5120 \\

~+DDSPO
& 0.8120
& 0.6660
& 0.7940
& 0.3070
& 0.3210
& 0.4060
& 0.5510 \\

\hline
\hline
\end{tabular}
}
\end{subtable}
\end{table}

\noindent\textbf{Evaluation on SD3.5-Large} \ \
To examine whether the benefit of DDSPO extends to a substantially
larger backbone, we apply DDSPO with TF-CPP to SD3.5-Large (8B).
We use 25K prompt pairs constructed from DiffusionDB prompts via random
token removal and fine-tune the model using LoRA with rank 128.
The model is trained in bfloat16 at a resolution of
$1024 \times 1024$ for 50 steps, with an effective batch size of 2,048
and $\beta=2{,}000$.
The reported results correspond to the checkpoint at step 50.

As shown in \cref{tab:supp_sd35_large}, DDSPO improves all GenEval
subcategories, increasing the overall score from 0.717 to 0.744.
The largest improvements are observed in color attribution
($0.565\rightarrow0.633$) and position
($0.263\rightarrow0.305$).
DDSPO also improves all T2I-CompBench subcategories, increasing the
overall score from 0.512 to 0.551, with particularly large improvements
in shape ($0.594\rightarrow0.666$) and texture
($0.723\rightarrow0.794$).
These results demonstrate that CPP-derived stepwise supervision remains
effective on a substantially larger text-to-image backbone.

\begin{figure}[t]
    \centering
    \includegraphics[width=0.78\linewidth]{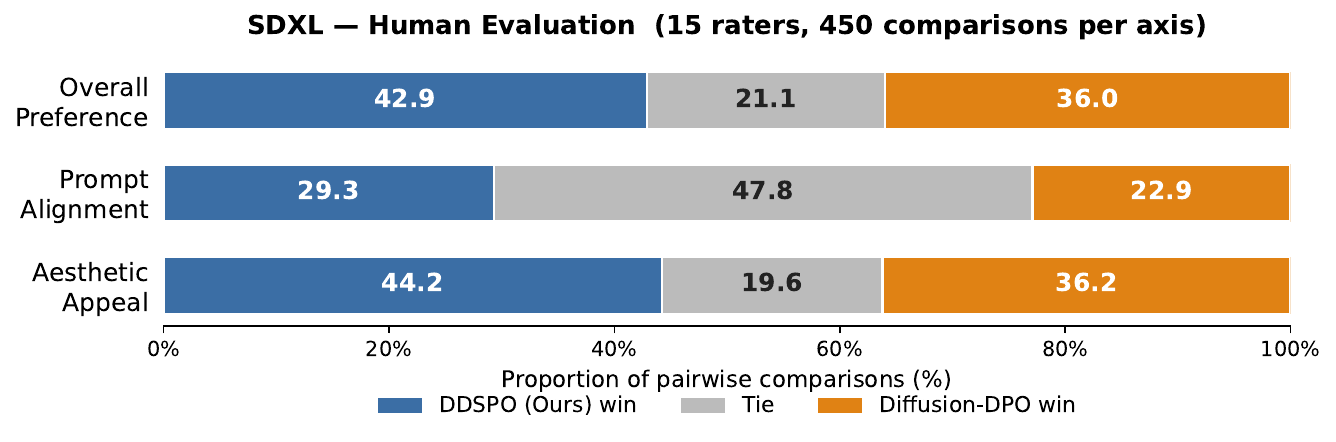}
    \caption{
    \textbf{Human evaluation on SDXL.}
    Fifteen annotators evaluate 30 prompt-matched image pairs,
    yielding 450 judgments per criterion.
    Each bar reports the raw proportions of preferences for DDSPO with
    DD-CPP, ties, and preferences for Diffusion-DPO.
    Ties are reported separately and are not redistributed between methods.
    }
    \label{fig:supp_human_eval}
\end{figure}
\noindent\textbf{Human Evaluation} \ \
We conduct a human evaluation on SDXL comparing DDSPO instantiated with
DD-CPP against Diffusion-DPO under matched Pick-a-Pic supervision.
Fifteen annotators evaluate 30 prompt-matched image pairs, resulting in
450 judgments for each criterion.
Images in each pair are generated using the same prompt and random seed. The model identities are hidden from annotators, and the left--right ordering of the two methods is randomized for each comparison.
Annotators assess overall preference, prompt alignment, and aesthetic
appeal, and may select either method or a tie.

As shown in \cref{fig:supp_human_eval}, DDSPO receives more preferences
than Diffusion-DPO across all three criteria.
For overall preference, DDSPO is selected in 42.9\% of judgments,
compared with 36.0\% for Diffusion-DPO.
For prompt alignment, the corresponding rates are 29.3\% and 22.9\%,
with a relatively high tie rate of 47.8\%.
For aesthetic appeal, DDSPO is preferred in 44.2\% of judgments,
compared with 36.2\% for Diffusion-DPO.
Overall, the study shows a consistent, although modest, human preference
for DDSPO.

\begin{table}[t]
\centering
\caption{
\textbf{Evaluation across five generation seeds.}
For the Diffusion-DPO and DDSPO with DD-CPP models compared in
\cref{tab:aesthetic_sota}, we report the mean and standard deviation
over five generation seeds.
}
\label{tab:supp_five_seed}

\scalebox{0.85}{
\begin{tabular}{llcc}
\hline
\hline
Backbone
& Metric
& Diffusion-DPO
& DDSPO \\
\hline

SD-1.5
& HPSv2
& $27.25 \pm 0.018$
& $\mathbf{27.59 \pm 0.013}$ \\

SD-1.5
& PickScore
& $21.35 \pm 0.010$
& $\mathbf{21.49 \pm 0.019}$ \\

\hline

SDXL
& HPSv2
& $28.55 \pm 0.008$
& $\mathbf{29.08 \pm 0.008}$ \\

SDXL
& PickScore
& $22.62 \pm 0.004$
& $\mathbf{22.71 \pm 0.004}$ \\

\hline
\hline
\end{tabular}
}
\end{table}

\noindent\textbf{Evaluation across Five Generation Seeds.} \ \
For the Diffusion-DPO and DDSPO with DD-CPP models compared in
\cref{tab:aesthetic_sota}, we report results over five random generation
seeds while keeping the model checkpoints, evaluation prompts, and all
other generation settings fixed.
As shown in \cref{tab:supp_five_seed}, DDSPO consistently improves the
mean performance across both backbones and both aesthetic-quality metrics,
while exhibiting only small variation across generation seeds.

On SD-1.5, DDSPO improves HPSv2 by 0.34 and PickScore by 0.14,
whereas the corresponding standard deviations are at most 0.018 and
0.019, respectively.
On SDXL, the HPSv2 improvement of 0.53 substantially exceeds the
standard deviation of 0.008, and the PickScore improvement of 0.09
similarly exceeds the standard deviation of 0.004.
Across all four settings, the performance gains are at least seven times
larger than the reported generation-seed variation, indicating that the
improvements are robust to the choice of generation seed.

\noindent\textbf{FID and Training--Evaluation Domain Shift} \ \
Our alignment models are trained on images generated from DiffusionDB
prompts, whereas FID is evaluated against the MS-COCO image distribution.
To quantify this training--evaluation mismatch, we additionally compute
FID between images generated from DiffusionDB prompts and MS-COCO
reference images under the same evaluation protocol.
This cross-dataset FID is 64.06, substantially higher than the FID of
13.05 obtained when images are generated from MS-COCO prompts and
evaluated against the same reference distribution.

This large discrepancy indicates that the FID increase after alignment
optimization is partly attributable to adaptation toward the
DiffusionDB-prompt distribution, rather than solely to a degradation in
image quality.
As reported in \cref{tab:align_comp_dpo}, all evaluated
preference-optimization methods increase FID relative to the base model
(13.05).
However, DDSPO yields the smallest increase, reaching 16.39
($+3.34$), compared with 18.02 ($+4.97$) for Diffusion-DPO,
18.28 ($+5.23$) for DSPO, and 17.12 ($+4.07$) for DSPO+CPP.
Thus, DDSPO achieves strong alignment improvements while inducing the
least distributional drift among the evaluated preference-optimization
methods.

\section{Comparisons with Alternative Approaches}
\label{sec:supp_additional_comparisons}

\begin{table}[t]
\centering
\caption{
\textbf{Applying SPO to text--image alignment.}
We compare the base SD-1.4 model with the model obtained by applying
SPO under our text--image alignment setting and report the full
category-wise results on GenEval and T2I-CompBench.
}
\label{tab:supp_spo_alignment}
\vspace{-1em}

\begin{subtable}[t]{\linewidth}
\centering
\caption{GenEval}

\scalebox{0.85}{
\begin{tabular}{l ccccccc}
\hline
\hline
      & \multicolumn{7}{c}{GenEval} \\
      \cmidrule(lr){2-8}
Model
& Single
& Two
& Counting
& Colors
& Position
& Color Attribution
& Overall \\
\hline

\rowcolor{gray!25}
SD-1.4
& 0.98
& 0.35
& 0.32
& 0.73
& 0.05
& 0.07
& 0.4245 \\

~+SPO
& 0.93
& 0.32
& 0.31
& 0.73
& 0.04
& 0.05
& 0.3953 \\

\hline
\hline
\end{tabular}
}
\end{subtable}

\vspace{0.5em}

\begin{subtable}[t]{\linewidth}
\centering
\caption{T2I-CompBench}

\scalebox{0.85}{
\begin{tabular}{l ccccccc}
\hline
\hline
      & \multicolumn{7}{c}{T2I-CompBench} \\
      \cmidrule(lr){2-8}
Model
& Color
& Shape
& Texture
& Spatial
& Non Spatial
& Complex (B-VQA)
& Overall \\
\hline

\rowcolor{gray!25}
SD-1.4
& 0.3593
& 0.3589
& 0.4024
& 0.1059
& 0.3096
& 0.3792
& 0.3192 \\

~+SPO
& 0.3982
& 0.3701
& 0.4080
& 0.1000
& 0.3113
& 0.3044
& 0.3153 \\

\hline
\hline
\end{tabular}
}
\end{subtable}
\end{table}

\noindent\textbf{SPO on Text--Image Alignment.} \ \
We additionally apply SPO under our text--image alignment setting using
SD-1.4 as the base model.
As shown in \cref{tab:supp_spo_alignment}, SPO does not improve the
overall alignment scores.
The overall GenEval score decreases from 0.4245 to 0.3953, with lower
performance in single-object generation, two-object composition,
counting, position, and color attribution.
On T2I-CompBench, SPO yields small improvements in color, shape, texture,
and non-spatial relationships, but these gains are offset by lower spatial
and complex-composition scores, resulting in a slight decrease in the
overall score from 0.3192 to 0.3153.

\begin{figure}[t]
    \centering
    \includegraphics[
        width=\linewidth
    ]{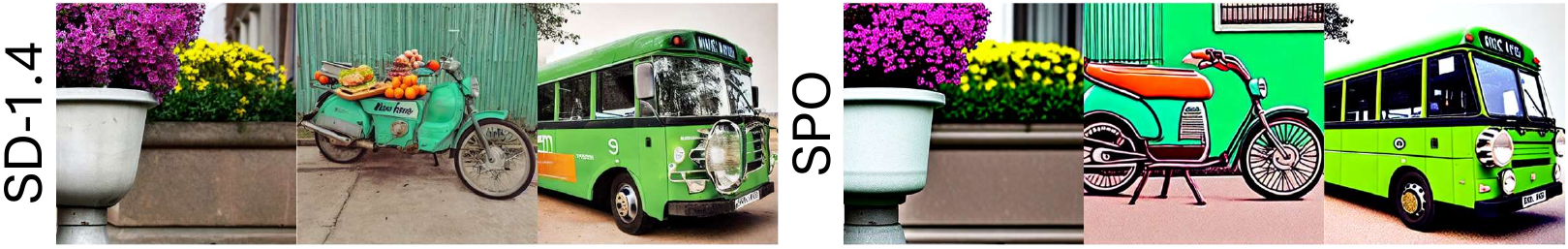}
    \caption{
    \textbf{Qualitative effects of SPO under the alignment setting.}
    SPO changes the visual appearance of generated images but does not
    consistently correct the semantic alignment errors in the base outputs.
    All images in each comparison use the same prompt and random seed.
    }
    \label{fig:spo_align}
\end{figure}

The qualitative examples in \cref{fig:spo_align} further suggest that
SPO mainly alters the visual appearance or style of generated images,
rather than consistently correcting their semantic correspondence with
the input prompts.

\begin{table}[t]
\centering
\caption{
\textbf{Comparison with SoftREPA on GenEval.}
We evaluate the publicly released SoftREPA checkpoint using our evaluation
pipeline and additionally provide the results reported in the original paper
for reference.
}
\label{tab:supp_softrepa}

\scalebox{0.85}{
\begin{tabular}{l ccccccc}
\hline
\hline
      & \multicolumn{7}{c}{GenEval} \\
      \cmidrule(lr){2-8}
Method
& Single
& Two
& Counting
& Colors
& Position
& Color Attribution
& Overall \\
\hline

SoftREPA (public checkpoint)
& 1.00
& 0.93
& 0.18
& 0.90
& 0.36
& 0.67
& 0.6740 \\

SoftREPA (reported)~\cite{lee2025aligning}
& 1.00
& 0.95
& 0.29
& 0.92
& 0.34
& 0.68
& 0.7000 \\

DDSPO (TF-CPP)
& 1.00
& 0.95
& 0.52
& 0.87
& 0.41
& 0.69
& 0.7384 \\

\hline
\hline
\end{tabular}
}
\end{table}

\noindent\textbf{Comparison with SoftREPA.} \ \
SoftREPA~\cite{lee2025aligning} is an annotation-free alignment method
that derives contrastive supervision from noisy in-batch samples.
For a controlled evaluation, we evaluate its publicly released checkpoint
using the same GenEval pipeline employed for our models.
As shown in \cref{tab:supp_softrepa}, the public SoftREPA checkpoint
obtains an overall GenEval score of 0.6740, while its original paper
reports 0.7000.
We include both results for transparency and use the public-checkpoint
result as the directly evaluated comparison under our protocol.

Under our evaluation pipeline, DDSPO with TF-CPP achieves an overall
score of 0.7384.
Compared with the public SoftREPA checkpoint, DDSPO performs better in
counting, two-object composition, position, and color attribution,
whereas SoftREPA performs better in the colors category.
Unlike SoftREPA's noisy in-batch contrastive supervision, DDSPO explicitly
constructs preferred and dispreferred transition-level targets from the
reference model conditioned on the original and degraded prompts.

\begin{table}[t]
\centering
\caption{
\textbf{Comparison with a CFG-style combined-target baseline.}
We test whether the effect of TF-CPP can be reproduced by directly
regressing toward a target constructed from the difference between
the original- and degraded-prompt predictions.
}
\label{tab:supp_cfg_direction}

\scalebox{0.85}{
\begin{tabular}{l ccccccc}
\hline
\hline
      & \multicolumn{7}{c}{GenEval} \\
      \cmidrule(lr){2-8}
Model
& Single
& Two
& Counting
& Colors
& Position
& Color Attribution
& Overall \\
\hline

\rowcolor{gray!25}
SD-1.4
& 0.98
& 0.35
& 0.32
& 0.73
& 0.05
& 0.07
& 0.4245 \\

~+CFG-style target regression
& 0.89
& 0.20
& 0.23
& 0.56
& 0.02
& 0.01
& 0.3196 \\

~+DDSPO
& 1.00
& 0.53
& 0.44
& 0.81
& 0.12
& 0.13
& 0.5045 \\

\hline
\hline
\end{tabular}
}
\end{table}

\noindent\textbf{Comparison with a CFG-Style Combined-Target Baseline.} \ \
Because TF-CPP uses the reference-model predictions conditioned on the
original prompt $\vc$ and a degraded prompt $\vc^-$, its training signal
may appear similar to directly optimizing along the difference between
these two predictions.
In particular, one could construct a CFG-style regression target by
moving away from the degraded-prompt prediction:
\[
\noise_{\mathrm{ref}}(\x_t,t,\vc)
+
\lambda
\left(
\noise_{\mathrm{ref}}(\x_t,t,\vc)
-
\noise_{\mathrm{ref}}(\x_t,t,\vc^-)
\right),
\]
where $\lambda$ controls the magnitude of this directional update.
To examine whether such a fixed target is sufficient to explain the
benefit of TF-CPP, we implement this combined-target regression as an
alternative baseline.

DDSPO with TF-CPP differs from this baseline in how the two reference
predictions enter the optimization objective.
It does not combine them into a single predetermined regression target.
Instead, TF-CPP provides the original- and degraded-prompt predictions
as separate preferred and dispreferred transition targets within the
DDSPO objective.
The original-prompt prediction acts as a positive anchor, while the
degraded-prompt prediction defines a negative direction through the
relative DPO-style comparison.
Thus, the strength of the negative update is determined adaptively by
the pairwise objective rather than by regressing toward a fixed linear
combination controlled by $\lambda$.

Under the same backbone, training data, optimization budget, and
evaluation protocol, the CFG-style combined-target baseline obtains
a GenEval score of 0.30, whereas DDSPO with TF-CPP achieves 0.50, as
shown in \cref{tab:supp_cfg_direction}.
This result indicates that the directional difference between the two
reference predictions alone does not account for the improvement;
retaining them as separate preferred and dispreferred targets within
the DDSPO objective is substantially more effective.

\section{Implementation Details}
\label{sec:supp_impl}
\noindent\textbf{Training Configuration} \ \
We fine-tune Stable Diffusion 1.4~\cite{sdm_v1.4},
1.5~\cite{sdm_v1.5}, SDXL~\cite{SDXL},
SD3-Medium~\cite{sd3}, and SANA~\cite{xie2024sana}. Stable Diffusion 1.4/1.5 are trained on 4$\times$ NVIDIA A100 GPUs, while SDXL, SD3-Medium, and SANA are trained on 2$\times$ NVIDIA H100 GPUs. Full fine-tuning is applied to Stable Diffusion and SDXL models in mixed precision (fp16). SANA is fine-tuned using LoRA~\cite{hu2022lora} in full precision (fp32) with rank 512, while SD3-Medium is fine-tuned using LoRA in mixed precision (fp16) with rank 128.

We follow the optimizer settings used in Diffusion-DPO: specifically, we use AdamW~\cite{adamw} for experiments with Stable Diffusion 1.5 and Adafactor~\cite{adafactor} for SDXL. For SANA, we use AdamW as the optimizer.
We use an effective batch size of 2048 pairs via gradient accumulation and train the main preference-optimization models for 100 steps. A linear warmup of 100 steps is applied to Stable Diffusion 1.4/1.5 and SDXL, while no warmup is used for SANA and SD3-Medium.
As in Diffusion-DPO~\cite{Wallace_2024_CVPR}, the learning rate is scaled with the divergence penalty coefficient $\beta$ as $\text{learning rate} = \beta / (2.048 \times 10^8)$. 


\begin{figure}[t]
    \centering
    \begin{tikzpicture}[xshift=+2em]
        \draw[line width=1.5pt, color=bkcolor] (-2em,0) -- ++(1em,0);
        \node[diamond, fill=bkcolor, scale=0.5] at (-0.8em,0) {};
        \node[font=\small, right=0.5em] at (0,0) {Compbench};

        \draw[line width=1.5pt, color=ourscolor] (7em,0) -- ++(1em,0);
        \node[circle, fill=ourscolor, scale=0.5] at (8.2em,0) {};
        \node[font=\small, right=0.5em] at (9em,0) {IS};
    \end{tikzpicture}
    \vspace{0.5em}

    \begin{subfigure}[b]{0.35\textwidth}
        \includegraphics[width=\textwidth]{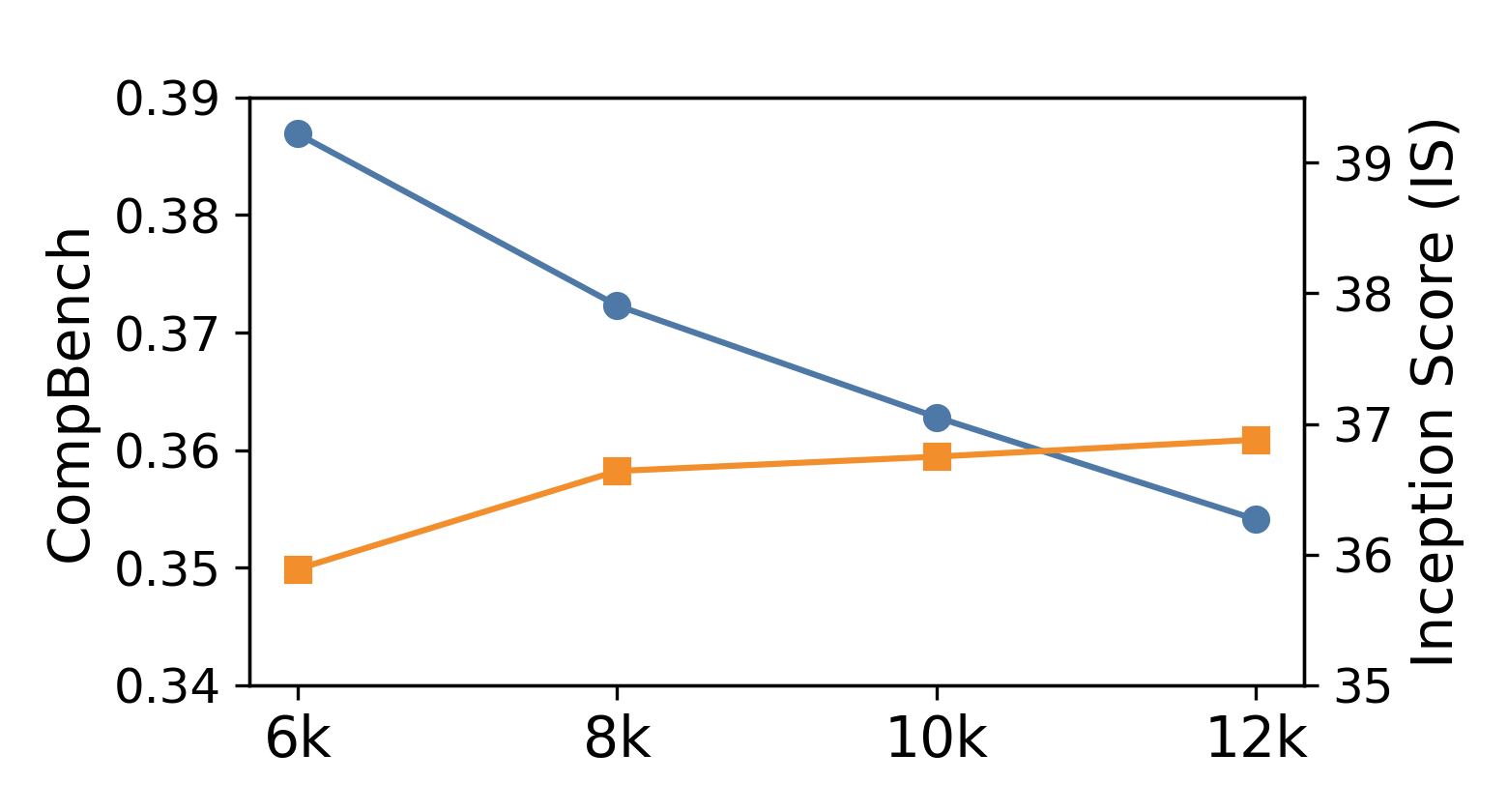}
        \caption{Diffusion DPO}
        \label{fig:sup:beta_D_DPO}
    \end{subfigure}
    \begin{subfigure}[b]{0.35\textwidth}
        \includegraphics[width=\textwidth]{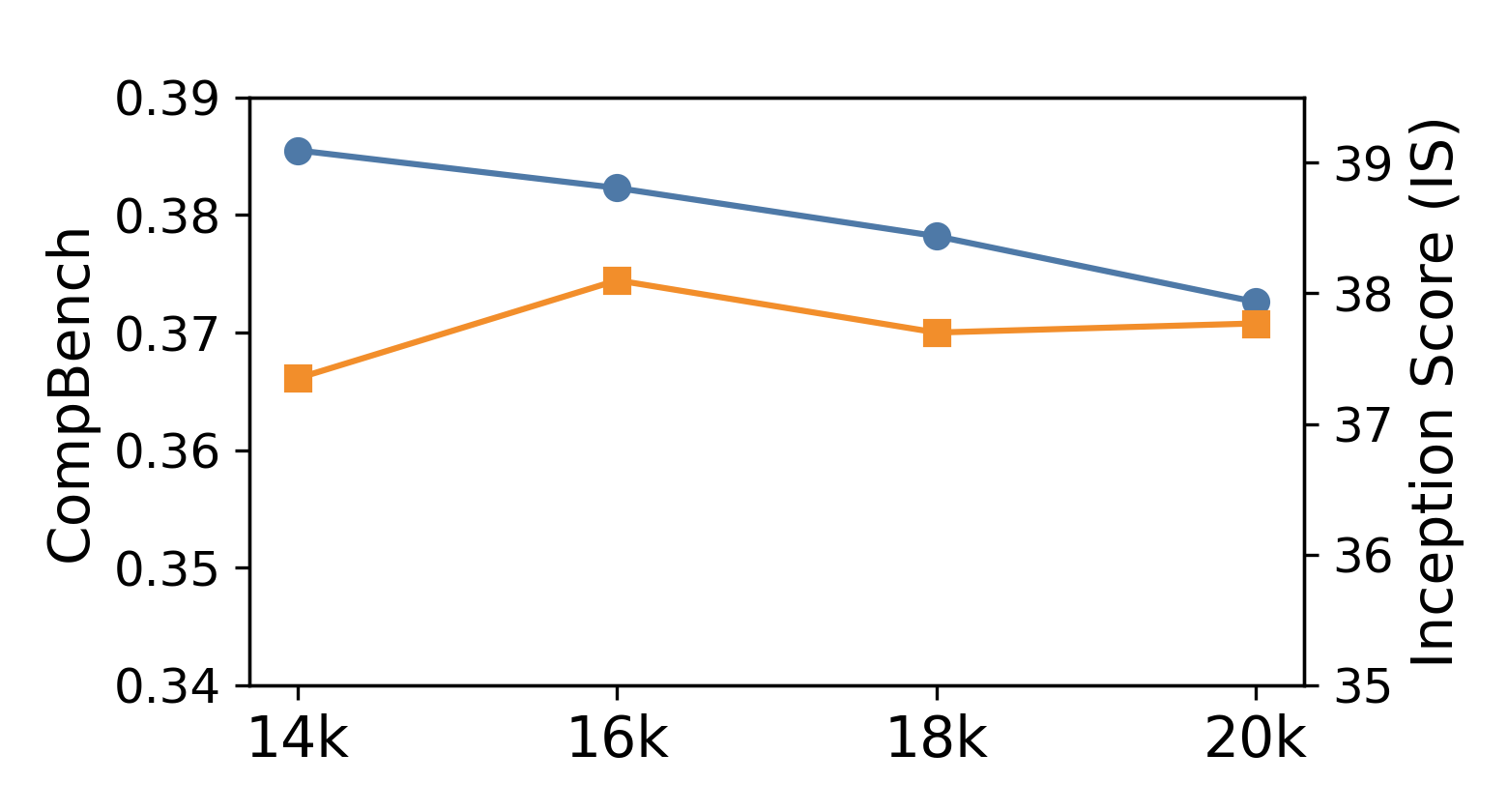}
        \caption{DDSPO}
        \label{fig:sup:beta_DDSPO}
    \end{subfigure}
    \caption{\textbf{Ablation Study on $\beta$.} Performance on Diffusion DPO (left) and DDSPO (right) with various $\beta$. The blue line denotes Compbench score while the orange line denotes IS. 
    }
    \label{fig:sup:beta}
    
\end{figure}
For TF-CPP, we sweep $\beta \in \{6000, 8000, 10000, 12000\}$ for Diffusion-DPO and $\beta \in \{14000, 16000, 18000, 20000\}$ for DDSPO on Stable Diffusion 1.4, as shown in \cref{fig:sup:beta}. The figure reports CompBench~\cite{huang2023t2i} and Inception Score (IS)~\cite{IS}, showing a trade-off between alignment performance and image quality: smaller $\beta$ values can improve alignment, but excessively small values often lead to noticeable degradation in image quality. Outside of these ranges, image fidelity deteriorates significantly for smaller $\beta$, while alignment gains diminish for larger values. Diffusion-DPO achieves the best balance at $\beta=8000$, while DDSPO consistently performs well at $\beta=16000$. We adopt these values as the default settings in the main experiments. The same value of $\beta=16000$ is used for SDXL, while $\beta=2000$ is used for SANA and SD3-Medium, both of which show stable performance.

For DD-CPP, we first train the contrastive policy pair on
Pick-a-Pic~\cite{pickscore} by learning separate winning and losing
policies. In this stage, we use LoRA with rank 4, a learning rate of
$1\times10^{-4}$, and an effective batch size of 2048. We train the
policy models for up to 300 steps and evaluate checkpoints every 50
steps, selecting the best checkpoint for each backbone. The selected
checkpoints correspond to step 150 for Stable Diffusion 1.5 and step
250 for SDXL. These policy pairs are then used to supervise the main
DDSPO training, where we set $\beta=8000$.

While our default settings may not represent the global optimum, we find
them to be robust across architectures and tasks. In our experiments,
using a larger $\beta$ combined with more training steps can match the
performance of the default $\beta=16000$, 100-step setup, but without
significant gains. Optimal configurations can differ not only from
Diffusion-DPO, due to differences in the loss formulation and supervision
signal, but also between DD-CPP and TF-CPP, since the quality and
structure of the contrastive policy pair depend on how it is instantiated.
In particular, TF-CPP relies on condition perturbation to construct
losing directions, while DD-CPP uses separately trained winning and
losing policies, which can lead to different regularization needs and
optimization behavior.

\noindent\textbf{Computational Cost.} \ \
We separately benchmark the training-time overhead of each objective
under a matched SD-1.5 setup using one NVIDIA RTX PRO 6000 Blackwell
GPU, bfloat16 precision, and a microbatch size of 16. Measurements are
averaged over 30 microbatches after five warmup iterations.
Diffusion-DPO and DSPO require 0.688 and 0.690 seconds per microbatch,
respectively, whereas DDSPO with TF-CPP requires 0.818 seconds,
corresponding to overheads of 18.9\% and 18.6\%, respectively. This
additional cost primarily arises from the extra reference-model
evaluation under the degraded prompt used to construct the dispreferred
transition target. The main DDSPO stage with DD-CPP requires 0.861
seconds per microbatch, which is 5.2\% slower than TF-CPP due to the
evaluation of the separately trained contrastive policies. Training the winning and losing policies together accounts for
approximately 30\% of the total DD-CPP training cost.

\noindent\textbf{Generation and Evaluation Settings} \ \
We use classifier-free guidance and DDIM sampling for all generations. For Stable Diffusion 1.4 and 1.5, we use a guidance scale of 7.5 and 25 sampling steps at a resolution of $512 \times 512$. For SDXL, we use a guidance scale of 5.0 and 25 steps at $1024 \times 1024$. For SANA, we use a guidance scale of 4.5 and 20 steps at $1024 \times 1024$. These settings are used for generating both $x_0^w$ and $x_0^l$ during training and evaluation.

Each model is paired with its respective default scheduler: PNDMScheduler~\cite{pndm} for Stable Diffusion 1.4 and 1.5, EulerDiscreteScheduler~\cite{eulerdiscretescheduler} for SDXL, and DPMSolverMultistepScheduler~\cite{dpmsolver,dpmsolver++} for SANA.

We report evaluation results using publicly released weights for all baselines, including Diffusion-DPO (trained on Pick-a-Pic), DPO-KTO, and Iterative Comparison. For CaPO, which does not release model weights, we report Geneval and CompBench results as stated in the original paper.

T2I-CompBench~\cite{huang2023t2i} evaluates different compositional aspects using specialized metrics: BLIP-VQA~\cite{li2022blip} is used for \textit{Color}, \textit{Shape}, and \textit{Texture}; UniDet~\cite{unidet} for \textit{Spatial}; and CLIP~\cite{clip} for \textit{Non-Spatial}. The default protocol for measuring \textit{Complex} is a 3-in-1 composite that combines BLIP-VQA, UniDet, and CLIP. However, CaPO~\cite{lee2025calibratedmultipreferenceoptimizationaligning} reports \textit{Complex} using only BLIP-VQA. To enable fair comparison with CaPO, we follow its evaluation setting in \cref{tab:align_comp_methods,tab:alignment_benchmark_results}, using BLIP-VQA alone to compute \textit{Complex}. For all other results, we adopt the original 3-in-1 composite as defined in CompBench, and use it to compute both the \textit{Complex} score and the overall average. For completeness, \cref{tab:alignment_benchmark_results} also includes the 3-in-1 version of \textit{Complex}.

\section{Limitation}
As with Diffusion-DPO, DDSPO requires careful selection of the balancing coefficient $\beta$, which controls the trade-off between reward maximization and divergence minimization. This reliance on hyperparameter tuning can add practical complexity, often necessitating task-specific calibration.

DDSPO also introduces additional computational overhead because its score-level supervision targets, $\noise_\star^{w}$ and $\noise_\star^{l}$, must be obtained from a contrastive policy pair during training. In the TF-CPP setting, this further requires sampling pseudo-preference pairs to construct the training data. In our SDXL setting, generating 200K pseudo-preference pairs for TF-CPP took approximately 5 H100-days, and DDSPO training for 100 steps required an additional 1 H100-day. While DDSPO achieves strong performance with only 200K pseudo-preference pairs and 100 training steps---substantially less data and computation than prior preference-optimization pipelines that often use 800K pairs and over 1,000 training steps---the cost of constructing and using contrastive policy pairs remains a practical limitation.

Finally, the overall effectiveness of DDSPO depends on the quality of the contrastive policy pair used to define the winning and losing directions. In this work, we instantiate it using DD-CPP and TF-CPP, but these may not be the optimal constructions. Developing more effective ways to obtain contrastive policy pairs---for example, through improved policy-pair training, stronger condition perturbation strategies, or alternative distillation-based constructions---could further improve DDSPO and remains an important direction for future work.

\section{Condition Perturbation Details for TF-CPP}
\label{sec:supp_perturbation}
To instantiate TF-CPP in text-to-image generation, we construct a degraded condition $\vc^-$ from each original prompt $\vc$ and use it to induce a losing direction through the pretrained reference model. For each prompt, the degraded condition is designed to yield lower-quality or misaligned images. We adopt different perturbation strategies depending on the supervision objective.

\noindent \textbf{Alignment Degradation (Random Removal)} \ \ 
For alignment supervision, we generate negative prompts by randomly removing 40--70\% of the tokens from the original prompt. This lightweight corruption introduces ambiguity and reduces semantic specificity, resulting in weaker text-to-image alignment.

\noindent \textbf{Alignment Degradation via LLM} \ \
We also leverage LLaMA3-8B~\cite{grattafiori2024llama3herdmodels} to generate semantically degraded prompts that obscure key visual concepts. As shown in \cref{fig:prompt_template_align}, the instruction guides the model to remove or generalize important elements, while preserving the overall structure of the original prompt. This allows us to obtain more compositional and realistic misalignment cases compared to random token deletion.

\noindent \textbf{Aesthetic Degradation via LLM} \ \
To supervise visual quality, we again use LLaMA3-8B to rewrite prompts in a way that intentionally reduces aesthetic appeal. The instruction template, shown in \cref{fig:prompt_template_aesthetic}, directs the model to degrade aspects like color, lighting, texture, and clarity while preserving the subject. This produces visually inferior prompts in a controlled and targeted way.

\begin{table}[t]
\centering
\caption{
\textbf{Effective degradation strength of different condition perturbations.}
We quantify the degradation induced by each strategy over 1,000 prompts
and relate it to the downstream performance reported in the main paper.
Lower CLIP image similarity and LAION aesthetic score indicate stronger
semantic and aesthetic degradation, respectively.
}
\label{tab:supp_degradation_strength}

\scalebox{0.85}{
\begin{tabular}{lccccc}
\hline
\hline
Perturbation
& CLIP Sim. $\downarrow$
& LAION Aes. $\downarrow$
& GenEval $\uparrow$
& HPSv2 $\uparrow$
& PickScore $\uparrow$ \\
\hline

Rand-removal
& \textbf{0.671}
& 5.371
& \textbf{0.5045}
& 27.28
& 21.36 \\

LLaMA (TA)
& 0.720
& 5.414
& 0.4854
& 27.25
& 21.36 \\

LLaMA (AQ)
& 0.749
& \textbf{5.248}
& 0.4758
& \textbf{27.51}
& \textbf{21.39} \\

\hline
\hline
\end{tabular}
}
\end{table}

\noindent\textbf{Effective Degradation Strength.} \ \
We quantify the effective degradation induced by each perturbation
strategy over 1,000 prompts.
For semantic degradation, we measure the CLIP image similarity between
generations conditioned on the original and perturbed prompts.
Rand-removal yields the lowest similarity of 0.671, compared with
0.720 for LLaMA (TA) and 0.749 for LLaMA (AQ), and also achieves the
highest GenEval score of 0.5045.
This result is consistent with stronger semantic degradation providing
a more informative negative direction for text--image alignment.
For aesthetic degradation, LLaMA (AQ) produces the lowest LAION
aesthetic score of 5.248, compared with 5.371 for Rand-removal and
5.414 for LLaMA (TA).
It also yields the best downstream aesthetic performance, reaching
27.51 on HPSv2 and 21.39 on PickScore.
These results suggest that task-specific perturbations that more
effectively degrade the target property tend to provide stronger
negative directions for TF-CPP.

\begin{figure}[t]
    \centering
    \includegraphics[width=0.85\linewidth]{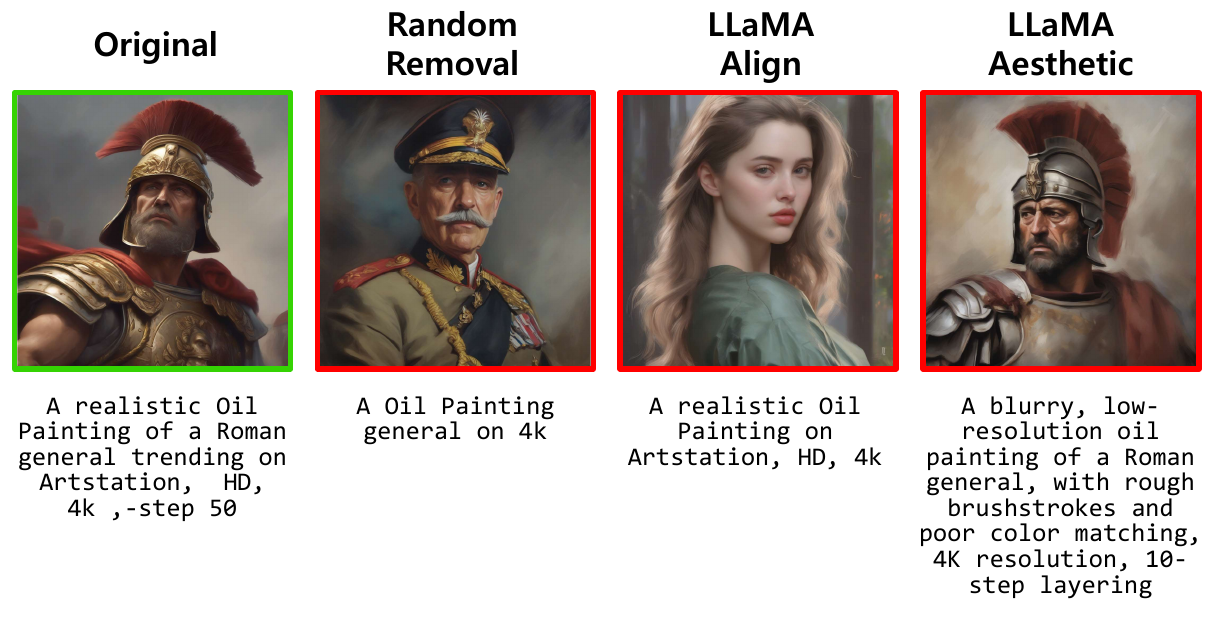}
    \caption{\textbf{Condition Perturbation Examples and Corresponding Generations.} Each column shows an original prompt \( c \) and its perturbed version \( c^- \) produced using different strategies—Random Removal, LLaMA-Align, and LLaMA-Aesthetic. Images are generated using SDXL for each prompt pair.}
    \label{fig:sup:prompt_pert_ex}
    \vspace{-0.5em}
\end{figure}

These perturbation techniques allow us to generate rich and scalable preference data without requiring manual annotations. Example prompt pairs are provided in \cref{fig:sup:prompt_pert_ex}.

\lstset{
  basicstyle=\ttfamily\scriptsize,
  breaklines=true,
  frame=single,
  backgroundcolor=\color{gray!10},
  captionpos=b,
  columns=fullflexible,
  keepspaces=true,
  showstringspaces=false
}

\begin{figure}[t]
\centering

\begin{subfigure}{0.95\textwidth}
\begin{lstlisting}[language=Python]
<|begin_of_text|>
<|start_header_id|>system<|end_header_id|>
You are a helpful assistant trained to rewrite image generation prompts into *compositional negative prompts*.

Your task is to generate a list of **negative prompts** that intentionally remove or generalize important elements from the original prompt.

Instructions:
1. Identify the **important words** (nouns or adjectives) central to the image composition.
2. Generate negative prompts that remove or alter these important words.
3. Keep the structure and style of the original prompt where possible.
4. If the prompt is too vague (e.g., "a macaw"), return an empty list.

Format your response as:
Important words: [...]
Final output:
{"neg_prompts": ["..."]}

No extra explanations or notes.
]

Below is the new prompt to process.
prompt: <input prompt>
<|eot_id|>
<|start_header_id|>assistant<|end_header_id|>
\end{lstlisting}
\caption{Instruction template for compositional degradation using LLaMA3-8B.}
\label{fig:prompt_template_align}
\end{subfigure}

\vspace{1em}

\begin{subfigure}{0.95\textwidth}
\begin{lstlisting}[language=Python]
<|begin_of_text|>
<|start_header_id|>system<|end_header_id|>
You are a helpful assistant trained to rewrite image generation prompts into two sets of outputs:

1. A list of **positive prompts** that enhance the aesthetic quality.
2. A list of **negative prompts** that degrade the aesthetic quality.

Instructions:
- Focus on aesthetic aspects like clarity, lighting, texture, resolution.
- Do NOT change the subject or meaning.
- For positive prompts, enhance descriptors.
- For negative prompts, reduce quality with terms like "blurry", "washed out", etc.

Format your response as:
Final output:
{
  "positive_prompts": ["..."],
  "neg_prompts": ["..."]
}

No extra explanations or notes.

Below is the new prompt to process.
prompt: <input prompt>
<|eot_id|>
<|start_header_id|>assistant<|end_header_id|>
\end{lstlisting}
\caption{Instruction template for aesthetic degradation using LLaMA3-8B.}
\label{fig:prompt_template_aesthetic}
\end{subfigure}

\vspace{-0.5em}
\caption{\textbf{Instruction Templates for Semantically Degrading Prompts.} Top: alignment-focused prompt perturbation using compositional degradation. Bottom: aesthetic-focused prompt perturbation targeting quality attributes. Both are used to construct pseudo-preference pairs for DDSPO training via LLaMA3-8B.}
\label{fig:prompt_templates_all}
\end{figure}

\section{Qualitative Results}
\begin{figure}[t]
    \centering
    \includegraphics[width=1\linewidth]{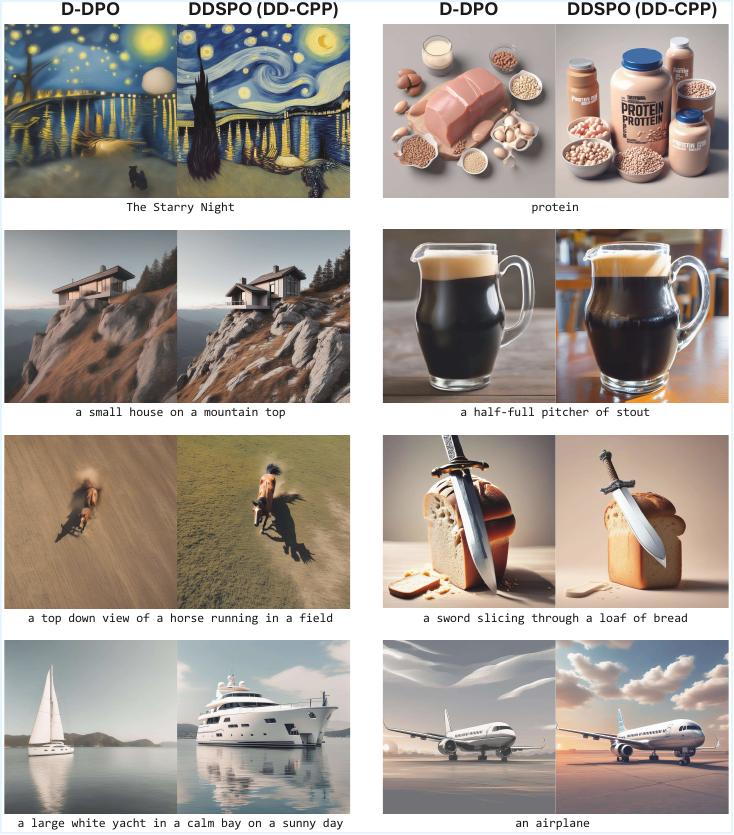}
    \caption{\textbf{Qualitative Aesthetic Comparison between D-DPO and DDSPO (DD-CPP).} Qualitative comparison of aesthetic results for D-DPO (left in each pair) and DDSPO with DD-CPP (right in each pair) on SDXL. Both methods are trained using Pick-a-Pic-based supervision.}
    \label{fig:sup_aesthetic_dpo_ddspo_ddcpp}
\end{figure}

\begin{figure}[t]
    \centering
    \includegraphics[width=0.972\linewidth]{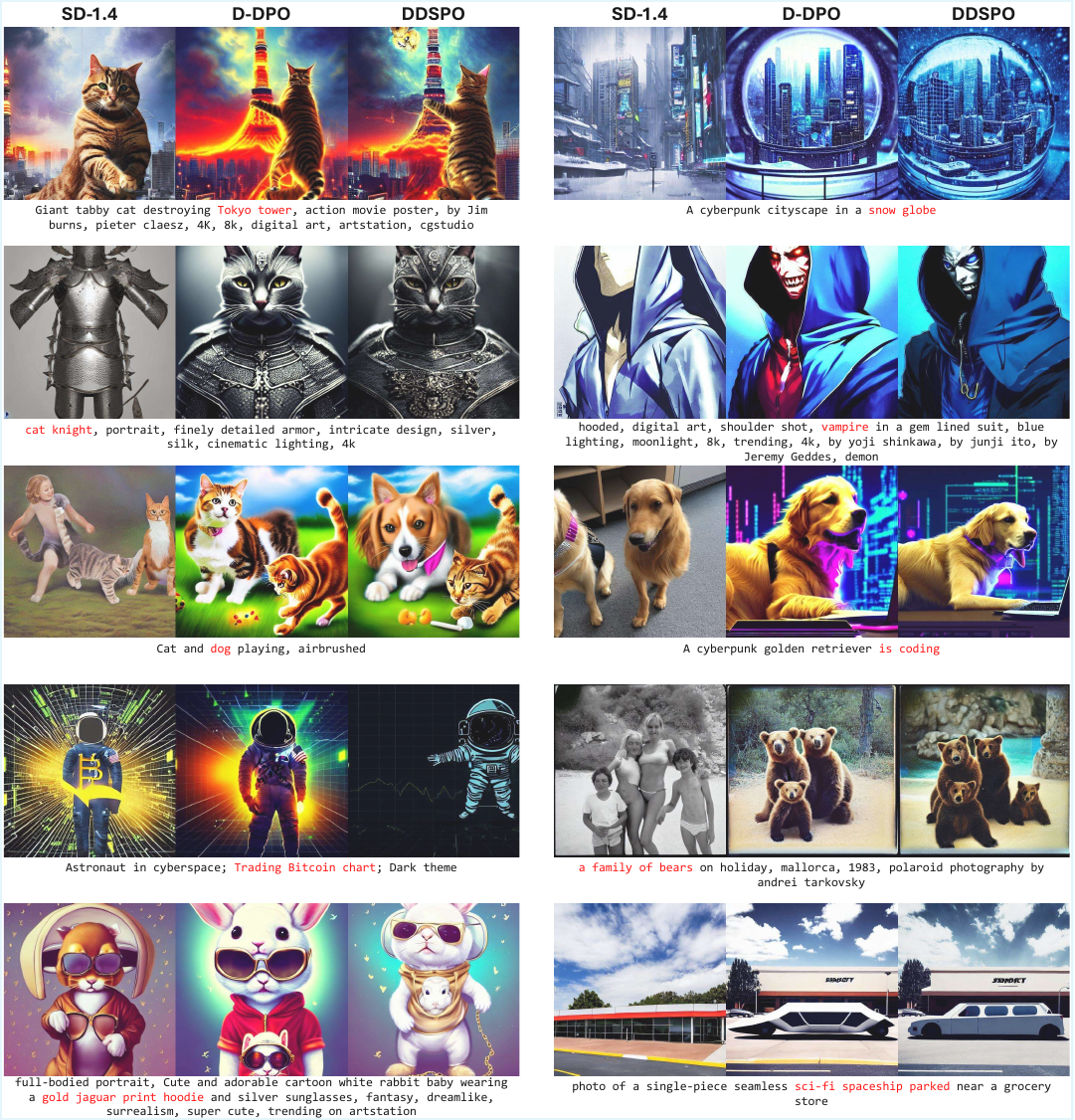}
    \caption{\textbf{Qualitative Alignment Comparison between SD-1.4, D-DPO, and DDSPO.} Qualitative comparison of alignment results for SD-1.4 (left in each pair), D-DPO (middle), and DDSPO (right). Both D-DPO and DDSPO are trained using TF-CPP-constructed supervision. Text highlighted in red indicates key objects or attributes missing from the SD-1.4 outputs.}
    \label{fig:sup_align_sd14}
\end{figure}

\begin{figure}[t]
    \centering
    \includegraphics[width=1\linewidth]{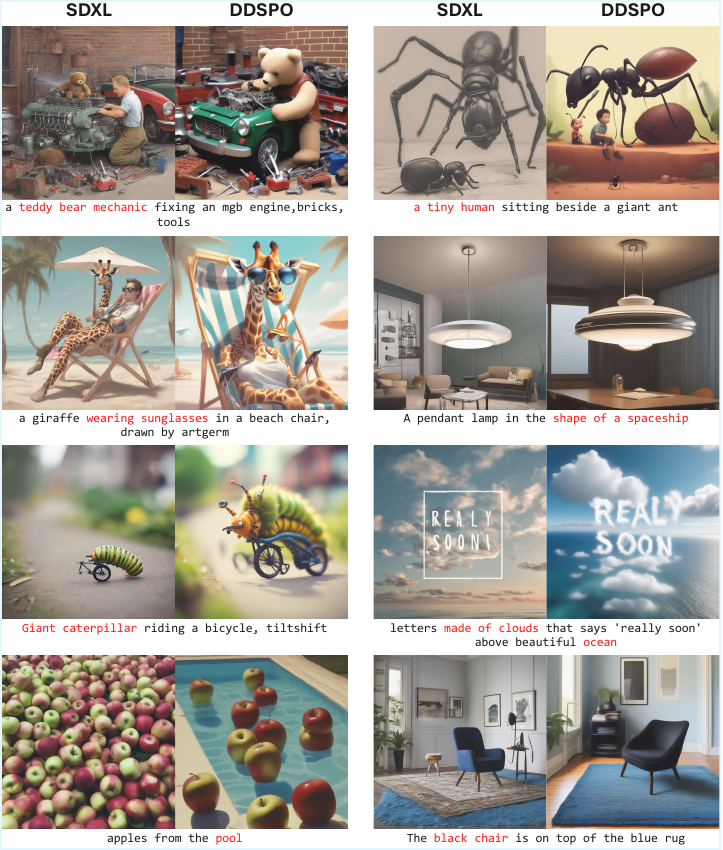}
    \caption{\textbf{Qualitative Alignment Comparison between SDXL and DDSPO.} Qualitative comparison of alignment results for SDXL (left in each pair) and DDSPO trained with TF-CPP (right in each pair). Text highlighted in red indicates key objects or attributes missing from the SDXL outputs.}
    \label{fig:sup_align_sdxl}
\end{figure}

\begin{figure}[t]
    \centering
    \includegraphics[width=1\linewidth]{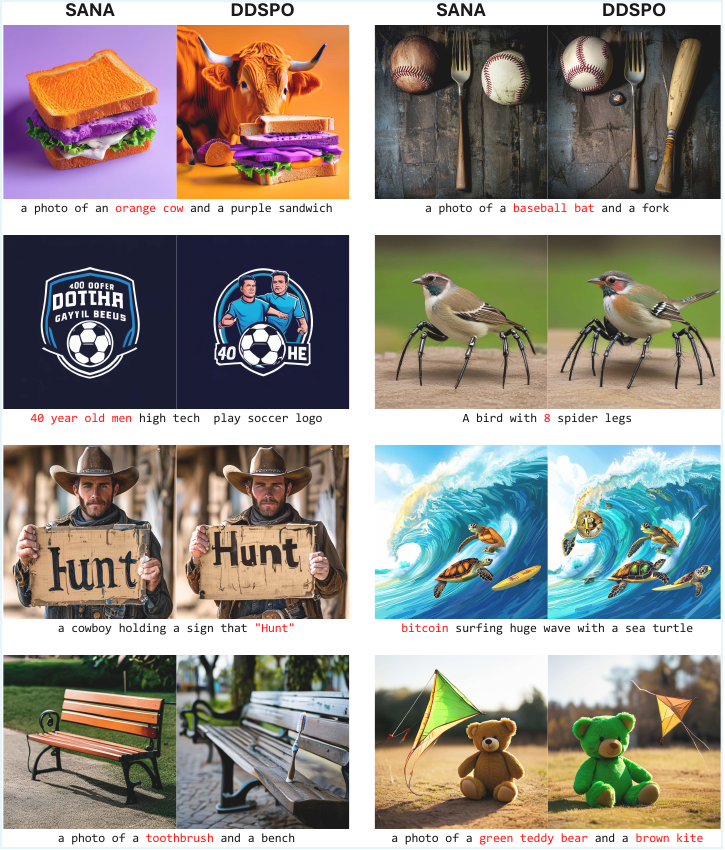}
    \caption{\textbf{Qualitative Alignment Comparison between SANA and DDSPO.} Qualitative comparison of alignment results for SANA (left in each pair) and DDSPO trained with TF-CPP (right in each pair). Text highlighted in red indicates key objects or attributes that were missing from the SANA outputs.}
    \label{fig:sup_align_sana}
\end{figure}

\begin{figure}[t]
    \centering
    \includegraphics[width=1\linewidth]{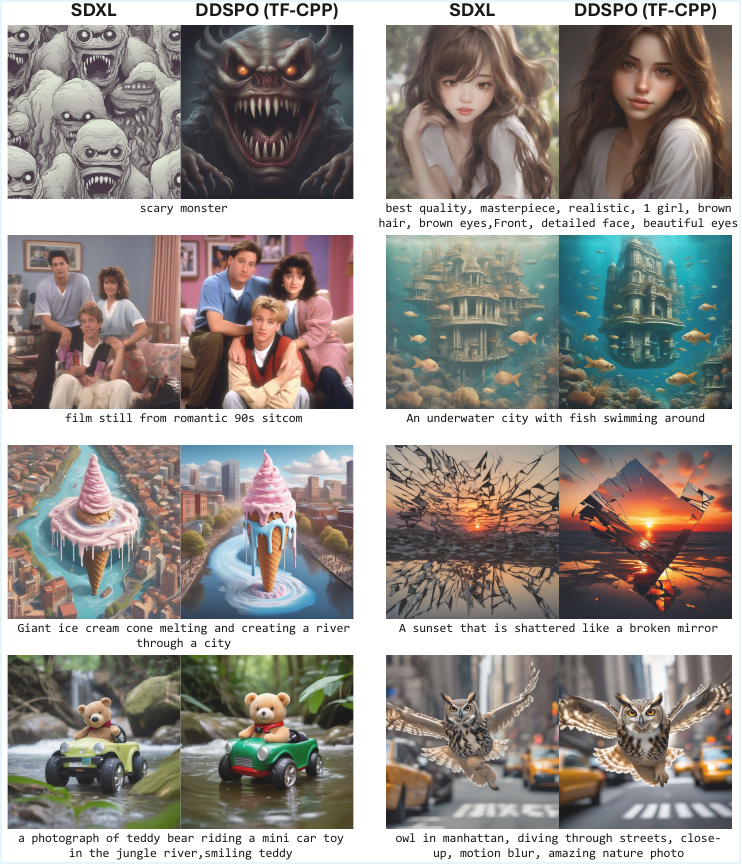}
    \caption{\textbf{Qualitative Aesthetic Comparison between SDXL and DDSPO~(TF-CPP).} Qualitative comparison of aesthetic results for SDXL (left in each pair) and DDSPO trained with TF-CPP (right in each pair) on selected prompts.}
    \label{fig:sup_aesthetic_sdxl}
\end{figure}

\begin{figure}[t]
    \centering
    \includegraphics[width=1\linewidth]{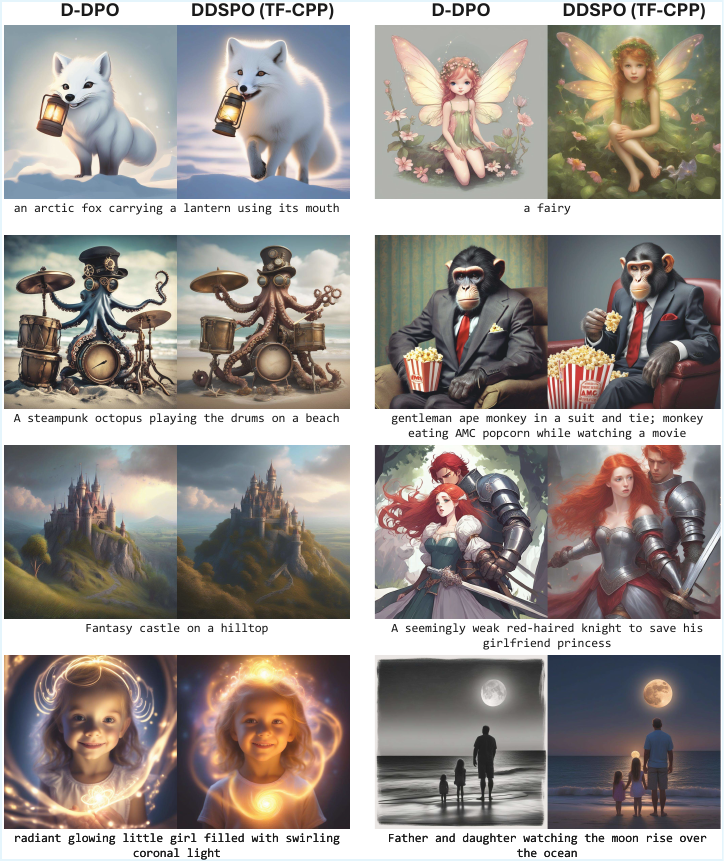}
    \caption{\textbf{Qualitative Aesthetic Comparison between D-DPO and DDSPO~(TF-CPP).} Qualitative comparison of aesthetic results for D-DPO (left in each pair) and DDSPO with TF-CPP (right in each pair). D-DPO is trained using human-annotated preference data from Pick-a-Pic, whereas DDSPO with TF-CPP achieves comparable or better aesthetic results without relying on human-annotated preference data.}
    \label{fig:sup_aesthetic_dpo_ddspo_tfcpp}
\end{figure}

We present qualitative comparisons under both aesthetic and alignment settings using SD-1.4, SDXL, and SANA. To ensure fair comparison, all images within each set are generated using the same random seed.
We first compare DDSPO against Diffusion DPO under matched settings. For aesthetics, \cref{fig:sup_aesthetic_dpo_ddspo_ddcpp} compares Diffusion DPO trained on Pick-a-Pic with DDSPO instantiated with DD-CPP, where both methods use Pick-a-Pic-based supervision. For alignment, \cref{fig:sup_align_sd14} shows results from SD-1.4 together with Diffusion DPO and DDSPO, where both preference optimization methods are trained using a TF-CPP-constructed dataset.
We then evaluate the generality of TF-CPP-based DDSPO across different base models. \cref{fig:sup_align_sdxl,fig:sup_align_sana} show that TF-CPP-based DDSPO consistently improves alignment over the corresponding base models across SDXL and SANA.
Finally, we present qualitative results showing that TF-CPP also enables data-free improvement in aesthetics. \cref{fig:sup_aesthetic_sdxl} compares TF-CPP-based DDSPO with its base model SDXL, and \cref{fig:sup_aesthetic_dpo_ddspo_tfcpp} compares TF-CPP-based DDSPO against Diffusion DPO, showing that DDSPO achieves competitive or stronger aesthetic improvements even without human-annotated preference data

\end{document}